\documentclass{article}
\usepackage[a4paper]{geometry}
\usepackage{cite}
\usepackage[pdftex]{graphicx}
\usepackage{amsfonts}

\usepackage{times}
\usepackage{url}
\usepackage{array}
\usepackage[font=footnotesize]{subfig}
\usepackage{graphicx}
\usepackage{mathtools}
\usepackage{enumitem}

\usepackage{alltt}

\def\boxx{{\vcenter{\vbox{\hrule height.3pt
          \hbox{\vrule width.3pt height6pt
          \kern6pt\vrule width.3pt}\hrule height.3pt}}\;}}

\def\impos{{\;\vcenter{\hbox{\rule{5mm}{0.2mm}}} \vcenter{\hbox{\rule{1.5mm}{1.5mm}}} \;}}

\def\lrarrow{\leftrightarrow \kern-8pt \rightarrow}

\def\2{\frac{1}{2}}

\def\grad{\nabla}

\def\beq{\begin{eqnarray}}
\def\eeq{\end{eqnarray}}
\def\2{\frac{1}{2}}

\newtheorem{example}{Example}

\begin{document}

\newcommand{\strust}[1]{\stackrel{\tau:#1}{\longrightarrow}}
\newcommand{\trust}[1]{\stackrel{#1}{{\rm\bf ~Trusts~}}}
\newcommand{\promise}[1]{\xrightarrow{#1}}
\newcommand{\revpromise}[1]{\xleftarrow{#1} }
\newcommand{\assoc}[1]{{\xrightharpoondown{#1}} }
\newcommand{\imposition}[1]{\stackrel{#1}{\impos}}
\newcommand{\scopepromise}[2]{\xrightarrow[#2]{#1}}
\newcommand{\handshake}[1]{\xleftrightarrow{#1} \kern-8pt \xrightarrow{} }
\newcommand{\cpromise}[1]{\stackrel{#1}{\frightarrow}}
\newcommand{\policy}{\stackrel{P}{\equiv}}
\newcommand{\field}[1]{\mathbf{#1}}
\def\R1{\mathbb{R}}
\newcommand{\bundle}[1]{\stackrel{#1}{\Longrightarrow}}

\title{Agent Semantics, Semantic Spacetime, and\\Graphical Reasoning}

\author{Mark Burgess}

\maketitle

\begin{abstract}
  Some formal aspects of the Semantic Spacetime graph model are
  presented, with reference to its use for directed knowledge
  representations and capacity for process modelling. A finite
  $\gamma(3,4)$ representation is defined to form a closed scaffolding
  set of operations that can scale to any degree of semantic
  complexity, with a highly simple rule set and a universal ontology.
  The Semantic Spacetime postulates bring predictability when
  reasoning, with minimal constraints on following pathways in graphs.
  The ubiquitous appearance of absorbing states in any partial graph
  means that certain graph processes leak information and represent
  entropy changing processes. The issue is closely associated with the
  issue of division by zero, which signals a loss of closure and the
  need for manual injection of remedial information.  The Semantic
  Spacetime model (and its Promise Theory) origins help to clarify how
  such absorbing states are associated with boundary information where
  intentionality can enter.
\end{abstract}

\tableofcontents


\section{Introduction} 

Semantic Spacetime (SST) is a discrete, graph theoretic `agent'
representation of configurations and process phenomena, used for
modelling scenarios that include knowledge representations, in the
form of labelled {\em directed
  graphs}\cite{spacetime1,spacetime2,spacetime3,cognitive}.  It
enables both qualitative and quantitative interpretations of processes
by combining physical and virtual concepts (from physics and
information science) into a Promise Theoretic agent
model\cite{promisebook}.  Promise Theory principles emphasize the autonomy or
{\em locality} of causal behaviour, so there are clear motivations for
modelling phenomena in this way.  As a graph theoretical structure, a
Semantic Spacetime is a collection of {\em nodes} (agents) joined by
{\em links} (channels for process information), both of which may have
annotations and numerical values associated with them.

A key application for Semantic Spacetime in artificial systems is to
represent `knowledge' (in its simplified sense) and process structures, such as those normally
associated with indexing methods or Semantic Webs, like the triple
store approaches of the Resource Description Framework
(RDF)\cite{rdf}. Further, one can explore how conceptual meanings and themes arise
from these ground states\cite{burgess2020testingquantitativespacetimehypothesis1,burgess2020testingquantitativespacetimehypothesis2}. These applications are the subject of many papers
spanning many decades, from the beginnings of taxonomies in the
biosciences to database schemas to tuplet spaces with supporting
ontologies. The persistent difficulties and logical inconsistencies of
these early methods, along with their lack of scalability, can be
traced back to the adoption of an entity-centric modelling of the
world in terms of `nouns' and `verbs' on a single scale---an
affectation that probably stems from modern philosophy. The use of
ontologies, in particular, to impose constraints on meaning at the
level of logic is fraught with rigidity and consistency issues,
because ontologies do not employ principles rooted in the processes of
the world.

Processes are sequences of steps, akin to algorithmic data
transformations, consisting of states unfolding over time as a set of `states' in the most
general sense.  Processes embody everything from manufacturing
(operations on raw materials) to simple arithmetic (operations on
numbers).  Our goal in representing knowledge is to model the changes
to every volume of spacetime we care about in terms of its role in a
cohesive semantic process that we can use as a model for reasoning.
Semantic Spacetime aims to be sufficiently formal to be precise and
yet sufficiently flexible to be widely useful.

\subsection{Spacetime characteristics}

A spacetime has certain regular concepts and `functional' behaviours
such as distance, time, scale, containment, direction: forwards,
backwards and so on.  In addition, other physical concepts such as
symmetry and order can be applied to spacetime `states' and the
operations involving them.

The concept of `functional semantics'\cite{watt1} is not a topic
normally associated with natural sciences like physics, except perhaps for a small number of category
theorists\cite{categories,quantumpictures}.  In linguistics, on the
other hand, semantics are closely allied with semiotics of
meaning\cite{langacker1,generativesemantics}. In Computer Science,
semantics have a more pragmatic meaning: related to the outcomes of
algorithms, i.e.  what behaviour or response is implied by certain
concepts and types. In addition to this, a meaning closer to the
linguistic one is used in connection with knowledge representations
and ontology\cite{ontologies}.

Although the terminology of `semantics' may be somewhat frowned upon
and do not apparently play a prominent role in discussions of natural
science, they are very much in evidence just below the surface in the
form of roles and representations. In the traditional mathematics of
rings, fields, and groups, for example, sometimes $x$ is a quantity of
something, with mundane semantics (what one would call a number), but
in other cases $x$ can be part of a generating function---an obstable
used as part of, say, a differential operator scheme for algorithmically for
representing certain patterns and process fragments, e.g. as in the
partition functions of thermodynamics and quantum field theory, or the ladder
operations of groups and graphs. Graphs therefore play a natural bridging role
between disciplines and we should include some discussion of how visual
imagery is used to give meaning to even abstract concepts and procedures.

Spacetime can certainly be described in terms of concepts then, but
here we are equally interested in the converse. If we turn this
association around, we can explore the way those concepts are perhaps
derived from spacetime concepts. After all, earlier in our
evolutionary journey there was nothing else to describe except where
things were, how to find them, and basic process impetus like
survival. There are likely insights that can be applied to other areas
like knowledge representation and reasoning.  The chief reason for
wanting to do this is the idea that human concepts ultimately derive
from concepts about space and time.
Consider the following phrases, each steeped in spacetime metaphors and yet apparently
representing quite different meanings:
\begin{quote}
Christmas is fast approaching.\\
We are getting close to an answer.We are approaching a result.\\
A long time coming. \\
The conclusion is just around the corner.\\
It's a far far better thing.\\
A sufficiently large contribution.\\
Time flies like an arrow.\\
The moment of truth approaches.\\
The subject is heavy reading.\\
Just north of 200 dollars.\\
Within the bounds of reason.\\
What makes it tick?\\
I would point out.\\
At this point of the proceedings.\\
It then took a left turn.\\
That's close enough to a win.\\
A long running affair.\\
On some level.\\
Just in time, within acceptable limits.\\
How long will you be?\\
We are moving ahead in spite of setbacks.\\
Our position is clear.
\end{quote}
These are examples of how we use spacetime semantics at the very root of
meaning. In addition, the gradual erosion of literal meanings and normalization of
metaphors that are later forgotten is a plausible model of how language
evolves in the first place\cite{unfolding,burgess2020testingquantitativespacetimehypothesis1,burgess2020testingquantitativespacetimehypothesis2}.
All this is a good reason to explore the idea that descriptions of events and the knowledge
associated with them can be simplified by appealing to the underlying spacetime types
and how they are used to express meaning.
It requires a more expansive and open minded approach to meaning
than is common in either physics or linguistics, yet this is what we
attempt to give a more formal substantial meaning to here.

\subsection{Euclidean vs graph}

The Euclidean-Cartesian view of space is the most prevalent in
science. Space is associated with translations in directions $x,y,z$
etc. The Minkowski space generalizes this to include coordinates for
an observer's local time too.  This view stems from the ballistic
origins of physics as a way of predicting the trajectories of weapons
fire.  Occasionally, one might use radial coordinates from a localized
point when considering perimeter boundaries, orbits, and Gaussian
enclosures and so forth.  However, the concept of translational
invariance of space is engineered so deeply into our tools that any
deviation from symmetry group thinking feels like a major deviation
from common sense. In the Newtonian tradition, we separate what happens
in space from the theatrical enclosure of abstract space itself.  

The classical view of spacetime attributes all phenomena to the
exterior space of points. Nothing happens `within' points.  Only in
the quantum theory do we encounter interior states, such as wave
functions, field values, spin, and other quantum numbers.  The idea of
hidden dimensions amounts to interior degrees of freedom at each
point. Together with agent models, quantum theory sees each location
in spacetime as having intrinsic or `interior' properties that may result
from unobservable processes within, but may also engage in more
classical cooperative processes between agents in an exterior ballistic view.

A graph $\Gamma = (N,L)$ is a pair of sets $N$ of nodes (also called
vertices) representing locations and $L$ of links (also called
edges). 
Graph oriented spacetime has rather different
challenges as compared to the more familiar continuum spacetime we are
taught in school.  In a continuum, everything is about the continuity.
The concept of infinity lies embedded in nearly every assumption,
without a clear sense of how. In a continuum, there is no reason for
one location (node) to point explicitly to another as in a graph. One
doesn't need a reason for the next point to be there--indeed, one
rather needs a reason why it might {\em not} be connected and ready to
receive some material body. There are many aspects of Newtonian age
physics that one takes for granted that fail to make certain sense on
closer examination. The continuum limit enables these slights of hand
in a way which is both elegant and audacious\cite{hesse}.

Arrows may be used arbitrarily in as labelling devices, but in the
world of causal events there are different kinds of graph. Petri Nets,
bigraphs, and Labelled Transition Systems are examples of attempts to
model process order as graphs, for instance.  On the other hand, the
standard Entity-Relation model of SQL databases uses arrows to
represent attributes, and uses counters to represent order.  In
Semantic Spacetime one makes no rules about this; instead, some simple
principles keep the meanings of arrows and attributes clear with a
weak implicit typing.

The relationship between graphs and continuum spaces in the literature
may lead to confusion. Embedding of a graph within a Euclidean space
is how one often gives meaning to concepts like distance and dimension
in graphs. The lattice approximations to Euclidean space uses regular
graphs with long range order to emulate Euclidean space with graphs.
Triangulation graphs can form approximate coverings of general
manifolds. These ideas are not how we are to understand Semantic Spacetime:
we are not trying to construct a skeletal embedding of trajectories in a Euclidean
theatre.

Similarly, we are not looking to yield a classification tree such as
a taxonomy, which is a hierarchical tree for organizing
membership of sets, subsets, collections, or grouping as a form of
containment. There, parent nodes branch out, forming a semantic branching
process\cite{grimmett1}.  The fact that all instances express a
conceptual similarity gives meaning to parent-child relationships.  A
taxonomy is an interior spanning tree for internal concepts.
Containment of things is a way of using a hierarchy
because partitioning of space follows a fractal branching rule.
Graphs therefore unify loosely topological and strictly geometrical
concepts. Semantic spacetime has aspects of both conceptual
decomposition and spanning of geometry. It is not a taxonomy or an ontology,
though we may find those embedded within it. It is neither a roadmap
embedded within Euclidean space with many redundant points in between.
One has only nodes and links with interior values. 

Embeddings interior and exterior to nodes are used widely in knowledge engineering.
This is how feature vectors are introduced as interior spaces in
artificial neural network models (see figure
\ref{etcspace})\cite{bishop1}.  In the world of ANNs, there are two
underlying spaces: the network of the ANN and the feature vector space
that is trained to represent concepts. This is almost the opposite of
the model of physics where Euclidean space is the embedding for
processes.  There are many attempts to formulate a mathematical
description of the dynamics of these representations, but they are
inscrutable property models\cite{selectivestate}.  In SST we take a
more symbolic algebraic approach to semantics, closer to language
encoding than a statistical model of meaning.

\subsection{Symmetries}

The role of symmetry is spacetime is to seek a minimalism of description
by emphasizing regular properties like homogeneity and isotropy.  Symmetry in Euclidean space
is represented by translation and rotation group transformations. A
discrete graph, on the other hand, has too little structure to allow
uniformity to form long range order. Nonetheless, there are processes
in graphs that render dissimilar actions symmetrical.  These include
process equivalence, redundancy of nodes and paths (also called
degeneracy in statistical physics), as well as the existence of `fixed
points' or absorbing regions of the graph that transmute any values
into a `zero' value. The semantics of how we deal with these are related
to the efforts to give meaning to division by zero in rings and fields\cite{bergstra6}
and we'll comment on these at the end.

Any kind of process from start to finish is basically isomorphic to
some kind of journey in some kind of space measure by some kind of
time.  Time in this sense is the Aristotelian concept of proper time as
countable changes, as observed by the agent concerned. These ideas
have been discussed at length in a variety of
works\cite{smartspacetime}.  The virtual or imagined space of concepts
has no obvious structure of its own.  It can be imagined in any shape
we like.
For example, any neural associative memory leads to the existence of
`hubs' or `appointed nodes' that join together many others in different
ways because of shared attributes or common containers. Without the
ability to express semantic symmetries, such as belonging to the same
class or type of information, it would be impossible to reduce the
entropy of graphs and create orderly information.

In any knowledge representation, which is neither complete nor
undirected graph, there will be hubs that lead to absorbing states and
information loss under certain transformations. The effect is to erase
information from the interior states of the nodes, which can only be
replaced with new boundary data from outside the graph, such as
outside policy choices.  Absorbing states are non-conserving of information.

\section{Semantics of spacetime and spacetime semantics}

We define the Semantic Spacetime model first in terms of the kinds of links
that connect nodes, and then in terms of the kinds of nodes that make sense
with those connections.

In a {\em process} view of the world, each point is a process event
that needs to be justified rather than postulated as a generalized
container for motion. This view is behind the famous relationship
between Feynman diagrams and the algebraic and differential generating
functional representations of Schwinger and Tomonaga in Quantum Field
Theory\cite{feynman1,dyson1,abbott1}. Further, both in Quantum Field
Theory and in Group Theory one has ladder stepping operators (creation
and annihilation) generating graphs by the action of an algebra of
operators over abstract states.  Underpinning all those variants are a
mathematical description of rings and fields that underpin everything
else practically as axioms.  Questioning these feels unproductive, yet
graph representations in particular have interpretations of arithmetic
that poke holes in the rules we take for granted.  One has to be clear
about whether returning to the same state is actually the same node in
a graph or a new version of a corresponding node which is similar but
distinguishable. These notions have been long understood in
statistical mechanics and thermodynamics\cite{reif1}.  In this regard,
a semantic spacetime is akin to causal
sets\cite{myrheim1,sorkin1,surya1}.

\subsection{The purpose of Semantic Spacetime}

Semantic Spacetime can be used in a number of areas
including agent modelling, collaboration networks like supply chains,
service mapping, formal reasoning, and narrative representations.  The
role of a knowledge representation is to capture experiences and to
abstract and generalize them, associate and collect them into
meaningful buckets that makes the knowledge easy to access and
interpret. For example, the flow of processes through a general graph
can represent:
\begin{itemize}
\item Scene description and forensic reconstruction.
\item Flow of utilities (e.g. electrical, water) through grids etc.
\item Communications networks like the telephone and Internet networks.
\item Dependency networks, like supply chains, ownership and market structures.
\end{itemize}

Any amount of detail can be added to a graph of meaning, in principle.
Recent developments in Large Language Models has placed much emphasis on
generating fluent language, which is a vast area of subtle meanings.
Before we get as far as semiotics and linguistic considerations, we need to look into
the constraints on meaning imposed by topology and conservation of information.
Dynamical considerations will always trump semantic considerations, because flow
is a dynamical phenomenon semantics
piggyback on quantitative amounts.

\begin{table*}[ht]
\begin{center}
\begin{tabular}{|c|c|c|c|}
\hline
\sc ST Type & \sc Forward & \sc Reverse & \sc Spacetime structure\\
\hline
\hline
&is close to & is close to &\sc PROXIMITY\\
\sc ST 0 = `$N$' & is similar to & is similar to&``near''\\
\sc `NEAR'     & sounds like & sounds like to&   Semantic symmetrization \\
     & is correlated with & is correlated with& similarity\\
\hline
\hline
     & enables & depends on & ordering\\
ST 1 = `$L$'&causes &is caused by  & \sc GRADIENT/DIRECTION \\
\sc `LEADS TO' & precedes & follows & ``follows''\\
              & to the left of & to the right of &\\
\hline
\hline
     &contains & is a part of / occupies& boundary perimeter\\
\sc ST 2 = `$C$' &surrounds  & inside & \sc AGGREGATE / MEMBERSHIP \\
\sc `CONTAINS' &generalizes & is an aspect of / exemplifies& ``contains'' / coarse graining\\
\hline
\hline
&has name or value &  is the value of property& qualitative attribute\\
\sc ST 3 = `$E$' & has property &  is a property of &\sc DISTINGUISHABILITY\\
\sc `EXPRESS & expresses attribute &  is an attribute expressed by & ``expresses''\\
\sc PROPERTY'& promises && Asymmetrizer\\
             & has approximation &approximates  & \\
\hline
\end{tabular}
\end{center}
\caption{\small Examples of the four irreducible association types, characterized by
  their spacetime origins, from \cite{spacetime3}. In a graph representation, `has attribute' and `contains'
  are clearly not independent, so implementation details can still compress the number of types.\label{assoc}}
\end{table*}

\subsection{The 4 relationship types}

The Semantic Spacetime model is basically a graph structure in which
nodes and links are used to represent and expose the compressed
meaning normally described in more fluent natural language.
This is in the spirit of formal languages and state machines; however, it
operates on a coarse level from the assumption that anything that happens is a process
than can be expressed in terms of some kind of spacetime journey.

The four types arise from the model of agents. In an agent model of
causality, which is fully localized, processes and information only
happen inside agents unless the agents cooperate to represent larger
structures. However, it is clear that activity only happens by agent
states changing. Ideas and structure cannot be imposed upon them from
outside without violating their local autonomy. This is a strong form
of locality as expressed in physics.  Based on the semantics of
agents, and the model of intent developed by Promise Theory, the
semantic spacetime model settles on four basic arrows between nodes in
a graph that are postulated to be sufficient for any semantic
description\cite{spacetime1,spacetime2,spacetime3}.  This remains a
hypothesis for now, but it is not a particularly original one.
Various authors have suggested that spacetime concepts underpin
natural language, on the basis that they are the only objective
concepts an organism has to bootstrap meaning from. Promise
Theoretically, one derives four kinds of promise or relation between
nodes in order to represent spacetime processes. These are called (see
table \ref{assoc}):
\begin{itemize}
\item $0$ {\sc near}: a directionless assertion of equivalence or proximity between two nodes
\item $\pm 1$ {\sc leads to}: a temporally or causally ordered arrow denoting a sequence of events
\item $\pm 2$ {\sc contains}: a spatially ordered collection of containment regions
\item $\pm 3$ {\sc expresses}: a locally promised attribute of a node or distinguishing mark
\end{itemize}
The directionless type 0 (proximity) may be interpreted as an
equivalence, approximately equal to, close to, etc. Type 1 (linear
sequential order) may represent time, or unidirectional ordering,
causation, dependency, etc. Type 2 (containment) may represent
membership in a group, generalization of a collection of concepts,
location inside or outside a perimeter, etc. 

See table \ref{assoc}, which originally appeared in earlier
papers\cite{spacetime3,cognitive} with some errors.
An alternative way of listing the 4 types is 
in a tabular form (see table \ref{tab1}).
Some of the relations address physical material attributes, while others represent virtual
or conceptual attributes used to describe scenarios.
\begin{itemize}
\item Physical is also assumed to refer to properties that are external to the agent, and are therefore `tangible*, `material', or
even `real'.
\item Virtual is assumed to refer to properties that are internal to
  the agent (or `somewhere else' in a hidden dimension), and are
  therefore intangible or purely informational.
\end{itemize}

\begin{table}[ht]
\begin{center}
\begin{tabular}{r|c|c}
  \sc DISTINCTION INFO  & \sc equivalence & \sc discrimination\\
\hline
& & \\
\sc (physical) situation  & \sc contains    & \sc leads-to \\
                      &   spatial         &  temporal \\
\hline
& & \\
\sc (virtual) information & \sc near        & \sc express property\\
                      & similarity/distance & attribute/name\\
\end{tabular}
\caption{\small A alternative tabulation of the arrow types as
  belonging to physical (exterior) and virtual (interior) realms.
  Expressions may either denote equivalence of neighbours or group
  roles, or discriminants of nodes or what forms
  sequences.\label{tab1}}
\end{center}
\end{table}

Any number of aliases or alternative interpretations of the four spatial
relationships are possible; indeed, they are encouraged in effective
communication for expressivity and qualification. However, the more
specialized kinds of arrow we introduce, the harder it is to reason about
them directly. 
However, we should also
be cautious that informal association of linguistic metaphor also
leads to confusions about the appropriate classification of meaning
under the irreducible types, as interpretation by metaphor is fluid in
human language\cite{unfolding}.
What the four types enable is a basic generic form of reasoning
about the meaning of a graph based on what kind of subject it describes.

\subsection{Resolving link type ambiguities with 3 types: {\em events}, {\em things}, and {\em concepts}}

\begin{table}
\begin{center}
\begin{tabular}{c|c|c|c}
\sc Role/meaning   &  \sc node type  & \sc properties & space/direction\\
\hline
&&\\
\sc activity/action   &   event  & ephemeral/realized  & timelike (process) agent \\
&&\\
\hline
&&\\
\sc subject/object   &   thing   & invariant/realized   & spacelike (snapshot) agent \\
&&\\
\hline
&&\\
\sc subject/object   &   concept & persistent/unrealized & virtual (role/intention) agent \\
&&\\
\end{tabular}
\caption{\small Agent node types in semantic spacetime. Virtual agents may be thought of as being
situated or `interior' attributes of spacelike or timelike agents.}
\end{center}
\end{table}

There is potentially some freedom in how to represent
information given the foregoing link types. Even with the four SST
categories there are some potential modelling ambiguities, so from a
computational perspective it's helpful to eliminate these with an
additional formality.  The residual ambiguity can be resolved by
recognizing nodes as belonging to one of three meta-types, denoted with small Latin letters
($e$,$t$,$c$), which are induced by the four link types.
The result is a generic $\gamma(3,4)$ representation, where:
\begin{itemize}
\item[$e$] {\em Events} ($e$) An event is a temporary or ephemeral
  phenomenon. Such happenings may persist or change in
  time, by {\sc Leads-to} which is a time like transition vector.
\item[$t$] {\em Things} ($t$) are persistent phenomena, physical or
  realized (manifest or reified) agents in what one would call
  classical space. A thing is a persistent phenomenon but may be
  created or destroyed.  They are agents that behave like matter.
\item[$c$] {\em Concepts} ($c$) A concept is an invariant notion which cannot be created or destroyed. It belongs to a virtual space of `unrealized' or `potential'
characteristics that can only be materialized by attaching to a
physical material agent. 
\end{itemize}

\subsection{The combined $\Gamma(N,L) \mapsto \gamma(3,4)$ representation}

With the selections described above, we have a compact set of organizing meta-types
that allow certain basic inferences to be made, relating to process semantics.
The typing immediate implies a few constraints that tighten up modelling.
We can summarize the design implications for these choices in a few rules.
\begin{enumerate}
\item Things may be contained but not expressed.
\item Concepts may be expressed but not contained.
\item Concepts become realized by anchoring them to things or events.
\item Verbs are dangling concepts without a subject or object to instantiate them.
\item Verbs that are anchored to subjects or objects (things) are events.
\item A state of being which is realized is an event   (john is/was happy, there was a moment of happiness).
\item A state of being which is unrealized is a concept (happiness).
\item A type of thing which is realized is a thing (a specific animal).
\item A type of thing which is unrealized is a concept (a general animal).
\end{enumerate}

\begin{example}[$e,t,c$ examples]
We shouldn't think of objects as having a single representation in a knowledge space. Different aspects of
a phenomenon have to be represented differently to obtain functional semantics:
\begin{itemize}
\item Mark's life is an event.
\item Mark' body is a thing.
\item Mark's character is a concept.
\item Mark's appearance is a series of events expressing observations of descriptive concepts.
\end{itemize}
\end{example}

\begin{example}[What is real?]
Which part of semantic spacetime does an idea belong to? Physical or virtual?
In the debates over the meaning of quantum physics, many authors
have questioned what `being real' means, e.g. whether we can say electrons
are real if they don't behave like large scale material bodies.
The use of `real' is too loaded with ideological connotations to be useful.
Adopting this terminology of $\gamma(3,4)$ avoids these ambiguous stumbling blocks.
One can be more precise about what is `real', `beable', or conceptual etc.
\end{example}

\begin{example}[Collectives or collective histories?]
Our ability to group events and concepts into a single namable entity
``all bicycles'' (a concept) from the evidence of many bicycle
instances (things) is powerful but misleading because it
exists only at a snapshot in time.  In general, a process
or history is a storyline, a conical structure of causal development
rather than an encapsulation of things. We only reduce it to a single
node in order to give it a proper name. Proper naming induces
``nodality'', but generalizations may be time dependent.
\end{example}
In the spacetime sense, things or concepts can be invariants. Concepts may be said to exist
yet be unrealized, i.e. they exist in the imagination, as a model or idea.  Events
are ephemeral but still realized by their happening.  The distinction between ideas that are
`realized' and `unrealized' is a simple clarification of the
implicit metaphorical indirection we use all too frequently in natural
language. Are we talking about the actual object or event $X$ or are
we talking about the idea of $X$ in some context, or all possible $X$?
Invariants can't be contained except by other invariants

\begin{example}[Many roles and contexts need different types]
Consider the word `library', which we might use quite liberally in language:
\begin{itemize}
\item A library is a concept.
\item The library is a thing.
\item Library is a role for a building.
\item Library is a location attribute for an event.
\item The opening of the library is an event.
\item A library contains books is a statement of concept.
\item The library contains books is a statement of things.
\item Library expresses attributes, old, well stocked, damp, at centre of city.
\end{itemize}
When there are multiple paths, the selection of a particular value in a multi-valued inverse
is affected by whatever semantics are applied to the arrows.
\end{example}

\subsection{Strategy in semantic graph representation}

Graphs as knowledge representations are typically understood either as classifications
or ordered concepts or as social networks:
\begin{itemize}
\item {Taxonomy}: a hierarchy of proper naming and sub-classification.
\item {Ontology}: a network of proper naming, classification and rules.
\item {Ad hoc property graphs}: an unstructured network of atomic items
and their relationships. All items are assumed to describe entities on a similar
scale or level.
\end{itemize}
In recent years, the proliferation of graph models and Semantic Web technologies has
encouraged users to model nodes in a graph by arbitrary linguistic atoms.
Any simple conceptual meanings are presumed `atomic'. 
The Semantic Web is best known example of a social network in the stigmergic sense,
with embedded order\cite{rdf,robertson1,fanizzi2}.  

Nodes are given meaning with supplementary `ontologies'\cite{ontologies}.
The ontology movement
is rooted in the primacy of first order logic for computer scientists.
The focus of an ontology is to specify and share meaning, whereas the
focus for a database schema is to describe data\cite{ontologydb}.  A
relational database schema has a single purpose: to structure a set of
instances for efficient storage and querying. The structure is
specified as tables and columns. An ontology can also be used to
structure a set of instances in a database. However, the instances
would usually be represented in a [possibly virtual] triple store, or
deductive database rather than directly in a relational database.
Ontology acts as a formal straitjacket because it only works if
everyone agrees to it.

\begin{figure}[ht]
\begin{center}
\includegraphics[width=11cm]{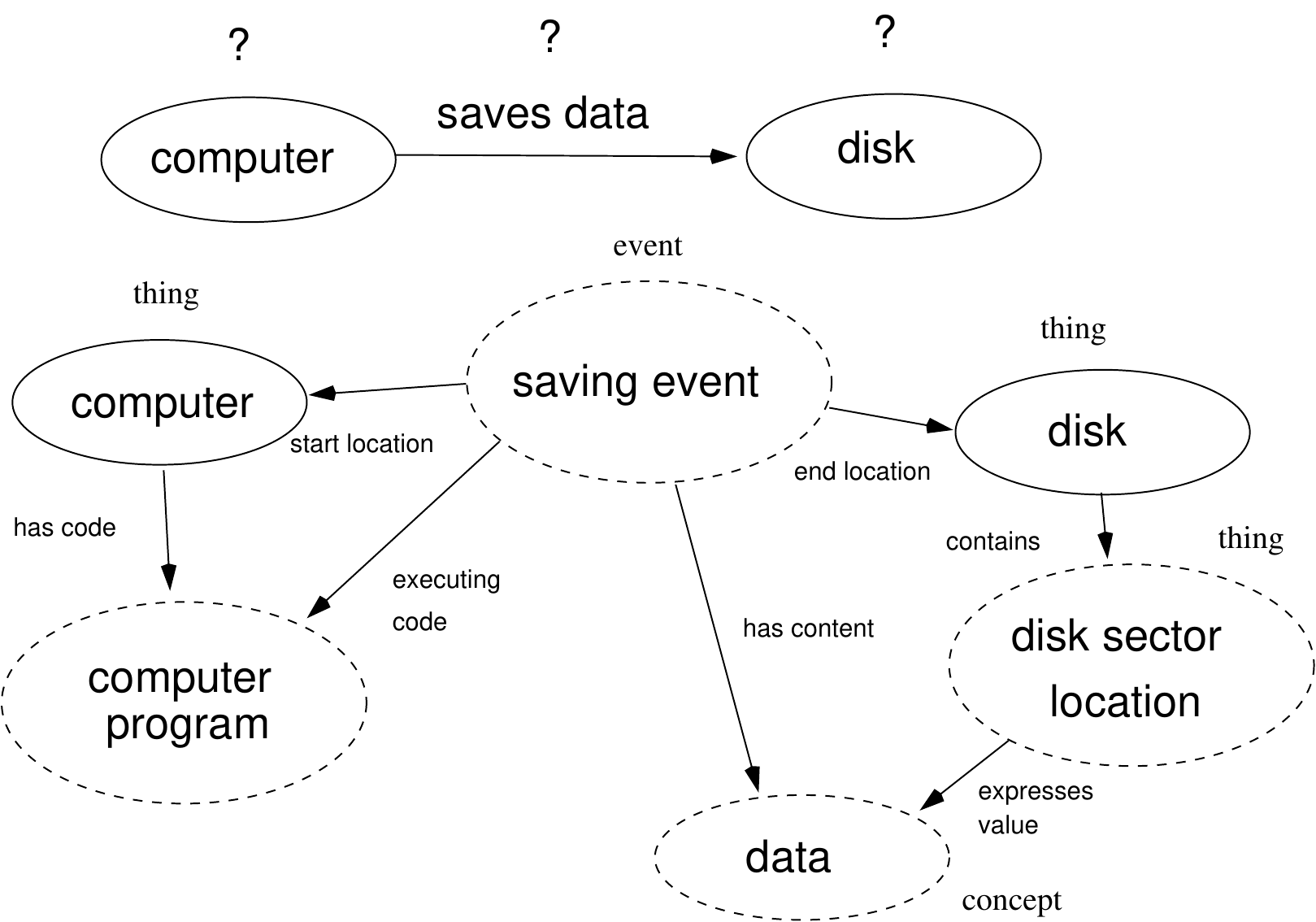}
\caption{\small A triplet graph compared to an $\gamma(3,4)$ graph.\label{graphcmp}}
\end{center}
\end{figure}

Trying to naively shoehorn natural language into a graph structure is
unproductive. For example, suppose we try to express
In common speech, we regularly transmute different aspects of a description into one another
all the time without much caution. Linguistic inference makes the algebraic underpinning of
natural language grammar difficult. The brain skips many pedantic typological
conversion steps, flattening out sequences as if coarse-graining steps by aggregation.
For example, consider:
\beq
\text{The computer}\text{ (saves data to) }\text{disk}.
\eeq
This is the kind of relation one often sees in naive triple store models.
The arrow `saves data to' is bursting with assumptions. It is so specific that it is
unusable outside of this example. In terms of the $\gamma(3,4)$ types we can compare it to
an SST graph (see figure \ref{graphcmp}).

One could argue that the `data' implicit in the upper link of figure
\ref{graphcmp} should be provided as an input to the graph.  Without a
decomposition, this can only be done by extending notion of links to
hyperlinks as in Milner's bi-graphs\cite{milnerbigraph,spacetime1}.
However, this introduces basic ambiguities that are easily resolved by
the $\gamma(3,4)$ decomposition.

The dotted lines indicate the subgraph representing the triple link. We see that it is not
a simple substitution.

If we say a thing smells of perfume we are adding an implicit
indirection from the concept of the smell of perfume to the liquid
perfume itself. A smell is not a liquid, but there is only only
natural interpretation of this muddling of proper names so no harm is
done.

\beq
\text{The cake} \text{ (looks like) } \text{mark's house}
\eeq

What type is ``looks like''? The cake is clearly a thing. Mark's house is a thing, but
``looks like'' expresses a similarity. If we write ``has the appearance of'' is looks
more like a concept attribute, which makes more sense. However now we have the rule
that only concepts (not things) can be expressed as attributes. Mark's house is not hanging
off the cake, only the likeness or image of mark's house, which is a conceptual representation.

These considerations are pedantic, like formal logic, because these are the typological
distinctions that lead to precision. But then why not simply use logic? We are trying to
avoid that level of specification by having only four types.

Concepts may express other concepts.
\beq
\text{Stubbornness} \text{ (means) } \text{explanatory concept}
\eeq
\begin{quote}
A hammer (may be used for) hammering event/activity\\
    "    (may be used for) hammering activity\\
An example of something is an event or instance not a member of a group\\
\end{quote}

How shall we decide the correct kind of link? When do we use express versus contains?
The meanings of these concepts are rooted in spacetime ideas.
\begin{quote}
London (makes) cars\\
Tractors (are supplied by) Massey Ferguson\\
Apples (have colour) green\\
Apple (is a) fruit
\end{quote}
In each of these cases, the kind of objects connected are wildly inhomogeneous.
Without semantic homogeneity there cannot be homogeneous reasoning. Every 
new arrow relation. It's not difficult to see that many implicit steps
are involved in each of these. Although we might say these abbreviated forms,
we imply many more steps. Clear the city of London does not make cars: some factory
located there contains a process of manufacturing and assembling. Making cars is a property
of the factory. However, as the container of the factory, the city somehow inherits that
promise of making cars. On the other hand, Sally is stubborn and Sally lives in London. This
does not mean that London is stubborn.

How can we resolve these idiosyncratic anomalies in the reasoning?
The key to unravelling these issues is to homogenize phenomena as generic processes.
What process transmutes cities into cars, machinery into companies, fruit types into colours, and
fruits into fruit types? The specific relations are not as important as whether something
is a chain of events, an explanation of an attribute, or a part of a larger whole.

If several nodes are joined by a common association, typically a property expression or group
container membership, then we can infer an implicit equivalence through correlation to the
third party. If we have 
\beq
A \text{ (is located at) } X\nonumber\\
B \text{ (is located at) } X
\eeq
we can assume that $A$ and $B$ are correlated somehow. However, this is likely lazy modelling.
Is this really an invariant fact, or merely an ephemeral meeting?
The database is describing a misleading snapshot of reality. Most likely $A$ and $B$ only
met in a single event of a more dynamical scenario.
$X$ should not be a location but an event.

\beq
\text{John Williams} \text{ (composed) } \text{Star Wars}
\eeq
The use of a verb `composed' lead to an ambiguous typing.
Is Star Wars the film, the soundtrack, the brand, the franchise, etc.
Is the composition a script, a piece of music, a clay model?
Human inference can make a good guess and derive meaning from this (perhaps not without
some confusion in general), but a machine algorithm relies on precise typing and cannot.

One could write explicitly
\beq
\text{John Williams} \text{ (composer has musical composition) } \text{Star Wars Soundtrack}
\eeq
This is then a reusable relation that implies two clear types:
composer and musical composition Written in this way as an invariant
factoid, it misses the opportunity to be precise.  We should, once
again, turn to events as specific instances (analogous to objects
instantiated from classes in computering).
\beq
 &\text{John Williams}&\text{composed the music for the movie Star Wars}\nonumber\\
   &"& \text{(event has composer) } \text{John Williams}\nonumber\\
   &"& \text{(event has composition) }  \text{Star Wars Soundtrack}
\eeq
If we write the link more explicitly:
\beq
 \text{Star Wars the movie (has soundtrack) Star Wars Soundtrack}
\eeq
we notice that the link itself induces types between the end points
as long as we choose link names accordingly. Thus the explicit logical
typing of objects is an unnecessary constraint as long as there is
broad consistency of naming. This simplifies the matter of search by random access
when we are unclear what we're looking for.

\begin{figure}[ht]
\begin{center}
\includegraphics[width=17cm]{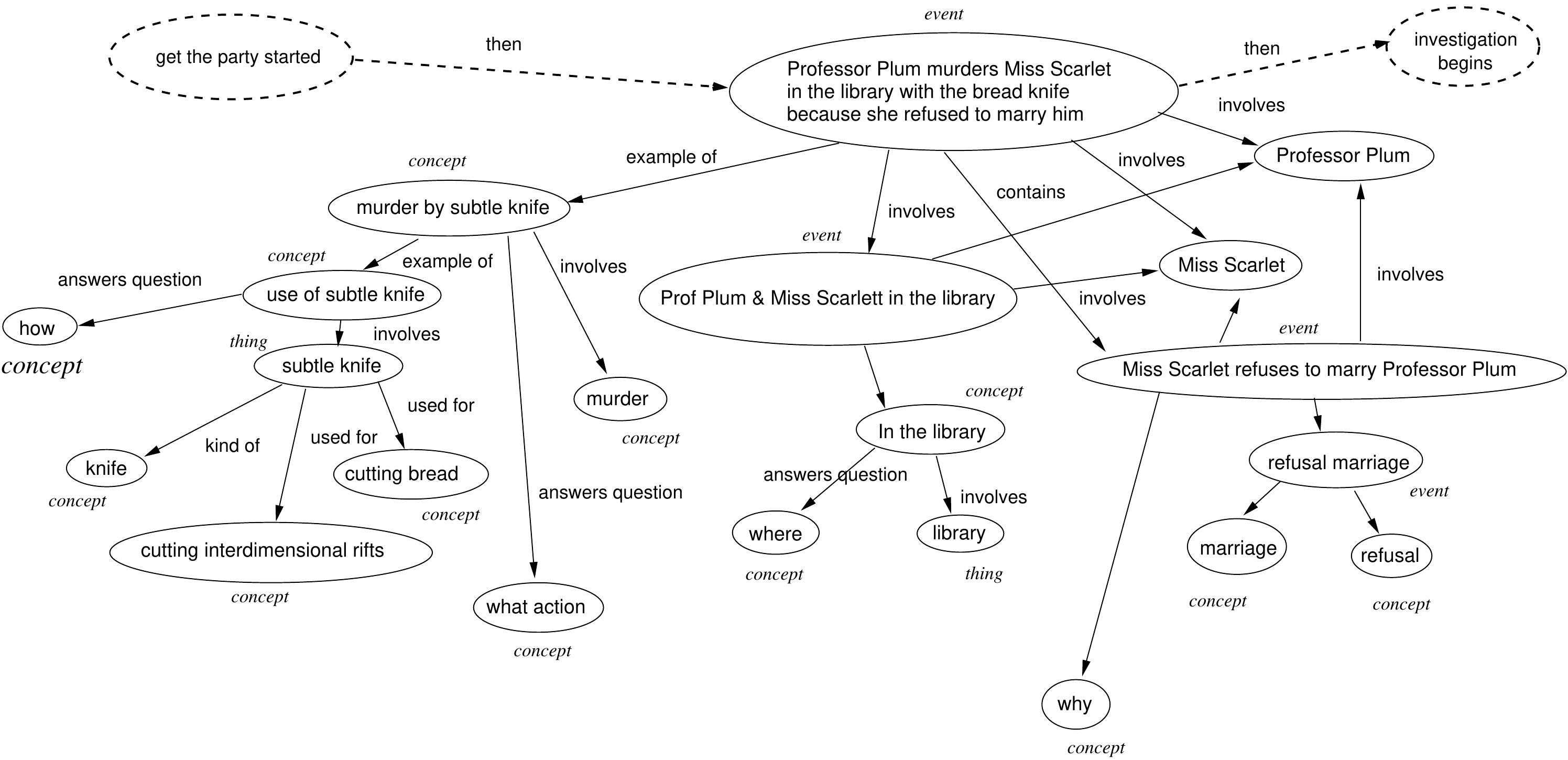}
\caption{\small Describing a more complicated scene in terms of pure spacetime semantics
is easier if we clearly describe events as more than the casual abbreviations
we use in language. This is analogous to the distinction between a class or datatype
in computer programming and a particular object instance of that class.\label{cluedo}}
\end{center}
\end{figure}

We use the name of an object and the concepts and events it is part of loosely in
natural language. Indeed, this is one of the strengths of natural language, which is
likely optimized for the implicit functional capabilities human brain: many such
connections can be left implicit and be filled in by inference. Part of the goal of
semantic spacetime is to help understand these matters.

\begin{figure}[ht]
\begin{center}
\includegraphics[width=10cm]{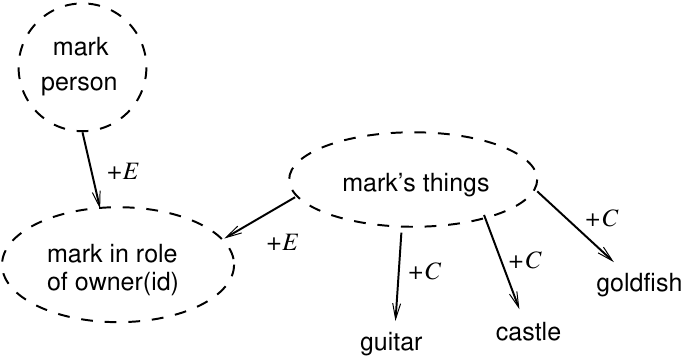}
\caption{\small We use the name of an object and the concepts and
  events it is part of loosely in natural language. Ownership is a
  mixture of concepts and things. The person `mark' is a thing so can
  only express a concept. To form a collection of things we need an
  entity that can contain them under a single umbrella. The estate of
  mark is thus a thing, and in order to express the ownership, we need
  to express the concept of ownership as a property of this
  collection. Since mark cannot (in a natural since) contain many
  other items that are possibly far away from mark, we use the
  attribute of connecting ownership as an expression of mark's
  identity (which is a concept). \label{ownership}}
\end{center}
\end{figure}

How we handle a complex issue like ownership may be seen in figure
\ref{ownership}.  Ownership is a mixture of concepts and things. The
person `mark' is a thing so can only express a concept. To form a
collection of things we need an entity that can contain them under a
single umbrella. The estate of mark is thus a thing, and in order to
express the ownership, we need to express the concept of ownership as
a property of this collection. Since mark cannot (in a natural since)
contain many other items that are possibly far away from mark, we use
the attribute of connecting ownership as an expression of mark's
identity (which is a concept).  Although this seems like a long winded
way of expressing the concept, it has the virtue of adhering to the
simple rules of the algebra, and removing the ambiguities about how to
define the concept of ownership. Ownership is decoupled from the
constraint of being a physical part of something else, but it still
can be.

\subsection{Directional  inference in the $\gamma(3,4)$ representation}

Spacetime operations are often associated with group transformations
that form chains or closures.  With a concept of graphical scale, and
the type separations of $\gamma(3,4)$, we can now attempt to make
certain inferences without any specific ontological rules.

The aspect of systems which is supported by Semantic Spacetime
is the frequently neglected dual importance of quantitative dynamics
with qualitative semantics\cite{burgess_search_2013}.  Process
representations that encode knowledge may have several
interpretations.  The most common understanding of computation is
based on the arithmetic conventions of rings and fields for addition
and multiplication of real numbers. Our universal adoption of standard
conventions means that we rarely question to consistency or meaning of
these rules, yet they are fraught with many ambiguities, which
normally only surface when faced with the `loose end of' division by
zero.

Our semantic model is relevant here too, as representations of
computations may involve either physical interlinking of agents (e.g.
agents collaborating on a chip die to perform a computation) or purely
informational maps recording conceptual models of the processes (e.g.
symbolic algebras).

The fact that we can represent computations as process graphs should
be no surprise.  It is a special case of expressing reasoning as a
graph, from ad hoc stories `once upon a time' to highly constrained
algebraic logics. The lesson, however, is that this is not as simple
as chaining together arbitrary triplets of named storage locations as
in an RDF graph.  The usual formalized notion of an ontology is of
little use too, since it is too specialized and once always needs to
extend or replace it. An ontology can never stabilize except by
limiting data or constraining the allowed phenomena.

The types $C,E$ do not `propagate' indefinitely in the sense that $L$ does. They are representative
of the snapshot one obtains by freezing temporal evolution in a fixed configuration.
We thus need to investigate the idea of scaling for inference. For instance, if we
take an example and generalize or specialize it (going `up' in the {\sc contains} or {\sc express} direction), 
does the generalization have the same properties as the example? For example:
\begin{itemize}
\item Mark is human.
\item Mark is tired.
\item All humans are tired.
\end{itemize}
This attempted syllogism is clearly wrong, but the opposite direction may be true.
\begin{itemize}
\item All humans are annoying.
\item Mark is human.
\item Mark is annoying.
\end{itemize}

As pointed out by Couch\cite{stories,inferences}, the approximate
notion of ``might be true'' or possibility links is sometimes the best
one can do when reasoning, There is no precise logical one hopes with
formal ontology. For example:
\begin{itemize}
\item If $A$ contains $B$ and $A$ contains $C$, then $B$ and $C$ might be near one another
on the scale of $A$.

\item If an event $E$ involves $A$ and $B$, then $A$ must have been near $B$
on the occasion of that event.

\item  If $A$ contains a collection nodes $B_i$ should they inherit properties expressed
by $A$?

\item If a collection of nodes $B_i$ are embedded in $A \rightarrow B_i \rightarrow C$,
then all the $B_i$ are symmetrical or equivalent with respect to this particular
process and may be considered part of a single supernode with redundant elements.
\end{itemize}

Interpretations like these are involved in the processes of reasoning
by deduction, induction, and abduction, etc.
The application of these ideas to unlabelled probabilistic transitions
of Markov chains is the way current diffusion models of machine learning
attempt to `reason' \cite{bishop1}.

\section{Graph-algebraic Semantic Spacetime}

Based on the foregoing definition of nodes and links, we can say more about
the properties of these graphs.
A graph may be represented structurally by a number of
matrices, in particular the adjacency and incidence matrices, which
represent maps of locations and flow gradients for the process
concerned.

\subsection{Proper names as semantic coordinates}

In knowledge systems, such as taxonomies, one attempts to give a
unique name to concepts in a contextually appropriate way.  The name,
although intended as a true representation of its unique meaning is
often not unique in practice.  In a coordinate system covering a
region of space, in which one attempts to label distinct locations, a
proper name is a kind of semantic coordinate. Unlike numerical
coordinate systems, proper names are often multivalued, which leads
both to opportunities and problems when reasoning.
For a simple graph representation in which the characteristics of the node
are the name itself, name associations may be considered tautologies, since
our ability to make distinctions relies on there being
observable differences. 

The scaling of names to groups and regions is not altogether trivial,
however.  In statistical subjects, the importance of
node or entity {\em distinguishability} has long been known and was shown by Boltzmann to
be associated with the degree to which systems are able to exert a
causal influence (free energy and entropy concepts)\cite{reif1}.  This issue of
unique identity affects the way we model concepts and things in a
graph, particulary when using spacetime as a model.

The concept of proper name interacts with the $\gamma(3,4)$ types.
A graph node is associated with a distinct identity. The numerical
identifier of the node or semantic identifier may be associated with
its `proper name', i.e.  the collection of symbolic or numerical
attributes that are expressed within or outside the node, i.e. the
union of interior attributes $S$ and the set of links $\{E(n_i)\}$ of
type `{\sc express property}'.
In semantic spacetime, scaling allows us to take an entire book of text as a node
in a graph. The proper name of the node is the entire text of the book.
The node can be decomposed into smaller parts  in a variety of ways
to express the meaning of the entity.

\begin{figure}[ht]
\begin{center}
\includegraphics[width=11cm]{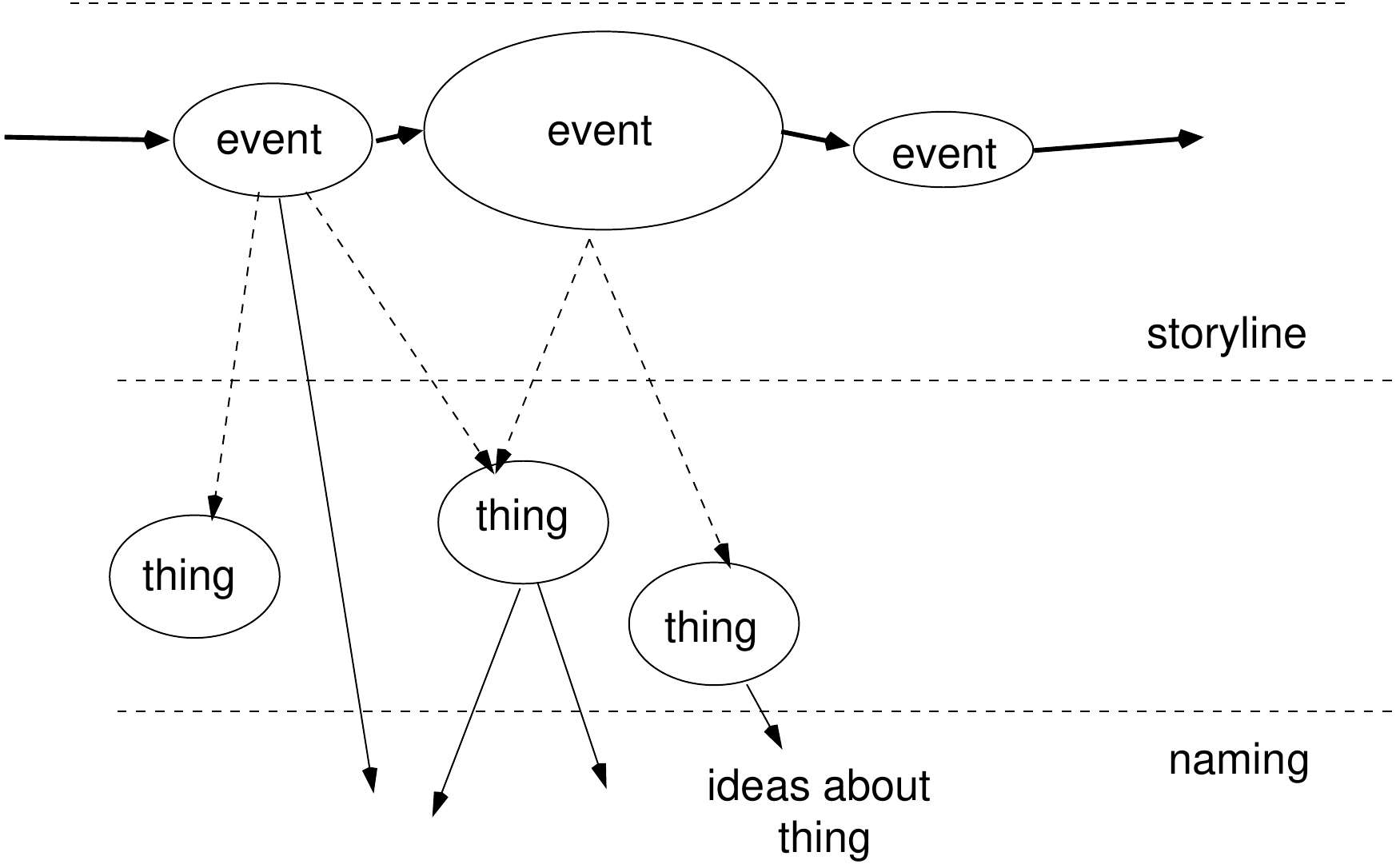}
\caption{\small The logical structure of events, things, and concept is subtle.
Events happen in real exterior space. Things exist in real exterior space, but ideas
about things are interior to the agents that express them. They can become shared
by interaction (like entanglement of quantum agents) but they are {\em a priori} private
on the interior of agents.\label{etcspace}}
\end{center}
\end{figure}

\begin{itemize}
\item A book, considered as an entity, has its entire text as its unique
identifier or proper name. We can give it several aliases, such as a
title, a cryptohash or an ISBN number.

\item The concept of the book  with its unique text is realized in many physical copies,
which express the concept of the book by reproduction. The physical books exists 
and can participate in different events  (see figure \ref{etcspace}).

\item Compressed descriptions of events, things, and concepts can be unified
under the umbrella of a name: `The Battle Of Britain', `Tractor',
`hunger'.  As patterns, we define names to be concepts or virtual
attributes, rather than physical realizations.  Names are usually
recognizable patterns, with a variety of manifestations: in speech or
writing, etc.  It's the content of the realizations that imbues the
name, not the mode of implementation.

\end{itemize}

\subsection{Graph structures: appointments and loops}

When several nodes point to a single node as their successor, we call
that an appointed node \cite{promisebook}.  This leads to a local
amplification of flow into the node.  
Nodes that are pointed to are also called `hubs', while nodes that
point to many are also called `authorities' in social graphs\cite{kleinberg99authoritative,graphpaper}.
An appointing node is the
equivalent predecessor or source for several successor nodes. This is
a division of the flow through the node into weighted distribution,
according to the link weights.  Appointing and appointed nodes
correlate their appointees implicitly, and thus form a common
dependency in reasoning.  Such nodes are important for several
reasons. Pragmatically, they a absorbing nodes and therefore lead to
division by zero issues.

Graphs may contain structures that have both semantic and dynamical consequences.
\begin{itemize}
\item {\em Sources and Sinks:} these are nodes that start and end a path through the graph. They exchange places if one changes the sign of the link type.

\item {\em Appointed nodes:} when several nodes point to a single hub that appointee is called an appointed agent in Promise Theory. The cluster of nodes all pointing / electing a single individual are thus correlated by the appointee (they have it in common). Such structures help us to see processes and process histories.
\end{itemize}

\begin{itemize}
\item For "leads to" arrows, these structures are confluences of arrows or explosions from a point.

\item For "contains" arrows, these structures are the containers or shared members

\item For "property expression" arrows, these structures are compositions of attributes or shared attributes common to several compositions

\item For "near" arrows, these structures are synonym / alias / or density clusters

\end{itemize}

Appointed nodes that are themselves appointed recursively form
nodes that are called `central'.

\begin{itemize}
\item A central node for `leads to' has a high level of involvement, implying a high mass for flows.

\item A central node for `contained by' arrows is the container.
\beq
\{n_i\} \promise{-\text{\sc contains}} n 
\eeq
\item A central node for `contains' is a member of several containers (a member of many categories).

\item A central node for expressing a property is a widely shared or common property.

\item A central node for originating properties has a rich spectrum of attributes.

\item A central node for being similar or near others implies a high
  density or a high level of redundancy in interpretation, perhaps a
  popular concept.
\end{itemize}

If we don't recognize these distinctions etc , the
four link classes or types run into ambiguities when trying to
classify the geometry of arbitrary semantics. For instance, {\sc express}
versus {\sc contains} are superficially similar, they both represent
interior states, however one refers to the physical materialized
makeup of a thing while the other refers to a concept.

\subsection{Propagation and terminating absorption of the $\gamma(3,4)$ types}

A node that propagates may form chains. Nodes that `absorb' arrows are
natural endpoints of graph flows.  Events are naturally propagating,
without any necessary end.  Both concepts and things are absorbing
types of node because there is a most primitive attribute in practice.

The mapping between directed graphs and sequences of events (as ST-1
`leads to' arrows) creates an obvious geometry for processes in a
graph. These arrows become associated with proper evolution of states,
e.g. Hamiltonian evolution in symplectic systems.

Consider examples for how each of the 4 types propagates.
\begin{itemize}
\item A terminating $L$ chain sequence is one in which there is no natural followup in the narrative:
\beq
\text{egg} \promise{\text{(gestates into)}} \text{caterpillar} \promise{\text{(becomes)}} \text{a butterfly} \promise{\text{(flies to)}} \text{tree}
\eeq
Notice that `a butterfly' is really the event of changing into the
state of being a butterfly, which is an event, since it wasn't a
butterfly before. Similarly `tree' is really a shorthand for `the
visitation of the butterfly to the tree', which depends on the
material thing `tree', but is not the tree. These language subtleties
trip modellers up frequently.  The event may refer to a thing, but it
is not the same as the thing. If we neglect to make these
distinctions, and believe too literally the meaning of our
abbreviations, we fall into the inconsistent the semantics of many
knowledge graphs. Leads to chains have no obvious end, unless we
choose to restrict them, e.g. to focus on a particular butterfly that
eventually dies or transforms back into its constituents. Ideally, we
would write the events more clearly (see figure \ref{cluedo}),

\item A $C$ chain:
\beq
\text{Mark} \promise{\text{(owns)}} \text{car}  \promise{\text{(made of)}} \text{atoms}  \promise{\text{(contains)}} \text{quarks} 
\eeq

\item An $E$ (or $P$) chain:
\beq
\text{diagram} \promise{\text{(has prop)}} \text{visual} \promise{\text{(has prop)}} \text{colour} \promise{\text{(has prop)}} \text{blue} \promise{\text{(has frequency)}} f  \promise{\text{(has units)}} \text{Hz} 
\eeq

\item An $N_t$ chain:
\beq
\text{Horsefly} \promise{\text{(looks like)}} \text{a butterfly} \promise{\text{(sometimes confused with)}} \text{a moth} \promise{\text{(resembles)}} \text{angel}.
\eeq
\end{itemize}
The types of things now enter into these chains of reasoning.

The metaphorical text of the song `Love is Like a Butterfly' can
clearly be written as a simple triplet: Love (is like) Butterfly.
However, this doesn't quite work in this more stringent
representation: \beq \text{Love} \not \promise{\text{(is like)}}
\text{a butterfly}, \eeq because love is a concept and butterfly is a
thing, and these cannot be alike while retaining this type
distinction.  Cleary natural language works by metaphor much of the
time, rather than by logic, and our brains are quite good at seeing
through these inferences.  This, on the other hand, makes the
challenge of more careful logical explanation from natural language a
non-trivial challenge.  Arguments may be presented for keeping or
eliminating metaphor in graphical knowledge representations. As long
as we want a more formal calculative framework for reasoning (which is
typically what we ask of machines), the shortcuts of natural language
are probably best avoided unless we can find a method for replacing
simplistic arrows with subgraphs that may be substituted in their place.
In a sense, this is what Artificial Neural Network representations are doing
on a probabilistic level.

\begin{figure}[ht]
\begin{center}
\includegraphics[width=4.5cm]{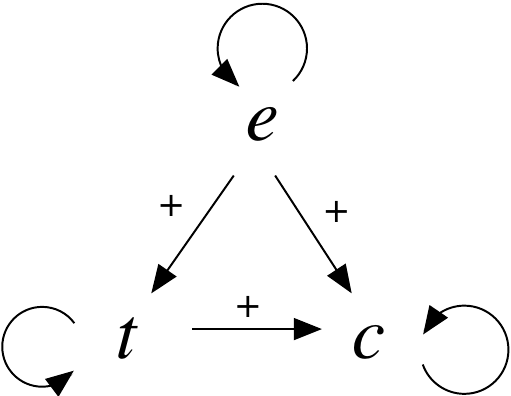}
\caption{\small Allowed semantic transitions between node types, by kinds of arrow.
There is a separation between virtual or conceptual states and physical or material
characteristics. Events are the encapsulating class for both of those.\label{sst2}}
\end{center}
\end{figure}

\subsection{Allowed type transitions of $\gamma(3,4)$}

We can now summarize the algebra of rules for valid graphs in the $\gamma(3,4)$ representation.
See figure \ref{sst2} and table \ref{tab2}.

\begin{table}[ht]
\begin{tabular}{c|l}
\hline
\sc Transition & \sc Explanation (example)\\
\hline
\hline
$e~(\pm L)~ e$ & An event can be followed by or lead to another event\\
$e~(\pm C)~ e$ & An event can contain or be part of another event\\
$e~(N_e)~   e$  & An event can be similar to another event by any criterion\\
$e~(+C)~    t$ & An event as a region of spacetime can contain a thing for its duration\\
$e~(+E)~    c$ & An event can express a property or concept (timestamp)\\
$e~(+E)~    c$ & can event can express a property or another event (celebration of Christmas of 74)\\
\hline
$t~(-C)~    e$ &  A thing can be part of an event, but an event cannot be part of a thing.\\
$t~(\pm C)~ t$ & A thing can contain or be part of another thing\\
$t~(+E)~    c$  & A thing can express a concept as an attribute (blue  car)\\
$t~(N_t)~ t$ & A thing can be close to or like another thing\\
\hline
$c~(+E)~ e$ & A concept can refer to an event as an attribute (that one time at band camp)\\
$c~(-E)~ e$ & A concept can be an attribute of an event (a time of happiness)\\
$c~(-E)~ t$ & Concepts can only be attributes expressed by things (blue car)\\
$c~(E)~ c$  & A concept can have properties or be a property of something else\\
   ~   & e.g. (blue is a colour)\\
$c~(N_c)~ c$ & A concept can be similar to another concept, (aquamarine,turquoise)\\
\hline
\end{tabular}
\caption{\small Explicit transitions allowed for events, things, and concepts through
the four link meta-types.\label{tab2}}
\end{table}
Due to linguistic inference, once again, it seems that an event ought to
be able to express or refer to another event as an attribute, e.g.
\beq
\text{Party celebrating the Olympics}\text{ (refers to) }\text{the Olympics}
\eeq
The Olympics was an event and the party is an event. However, if we seek
a clean distinction we need a more pedantic eye. The use of Olympics
in this case refers to the collective memory of the event, 
not the actual happening event itself--which is a concept without
physical manifestation. We might prefer to mention Olympics only once in
a knowledge graph, but then we would render all uses of the word equivalent,
which is semantic nonsense.
The same type of reasoning can be applied to explain why things cannot be expressed
as properties.
For concepts that seem to express things, e.g.
\beq
\text{fast food hatred}\text{ (is about) }\text{fast food},
\eeq
we observe that the reference to `fast food' is not a reference to an 
instance of fast food, but rather a reference to the whole class of things we
call fast food, which
is a concept.  Our minds quickly create short cuts through these
matters, yet we are also intuitively aware of the distinctions. This
underlines (as is well appreciated) that, without quite sophisticated automated analysis
capabilities, natural language could not be understood literally. Our penchant for
metaphor is busily at work in natural language\cite{unfolding}.

The consistency of these relations can be show using the matrix algebra in section \ref{matrices}.  In short, we can use this decomposition to define what are events, things, concepts.
One could add more detail, but we are trying to compress description into simple
elements rather than exfoliating.

\begin{figure}[ht]
\begin{center}
\includegraphics[width=4.5cm]{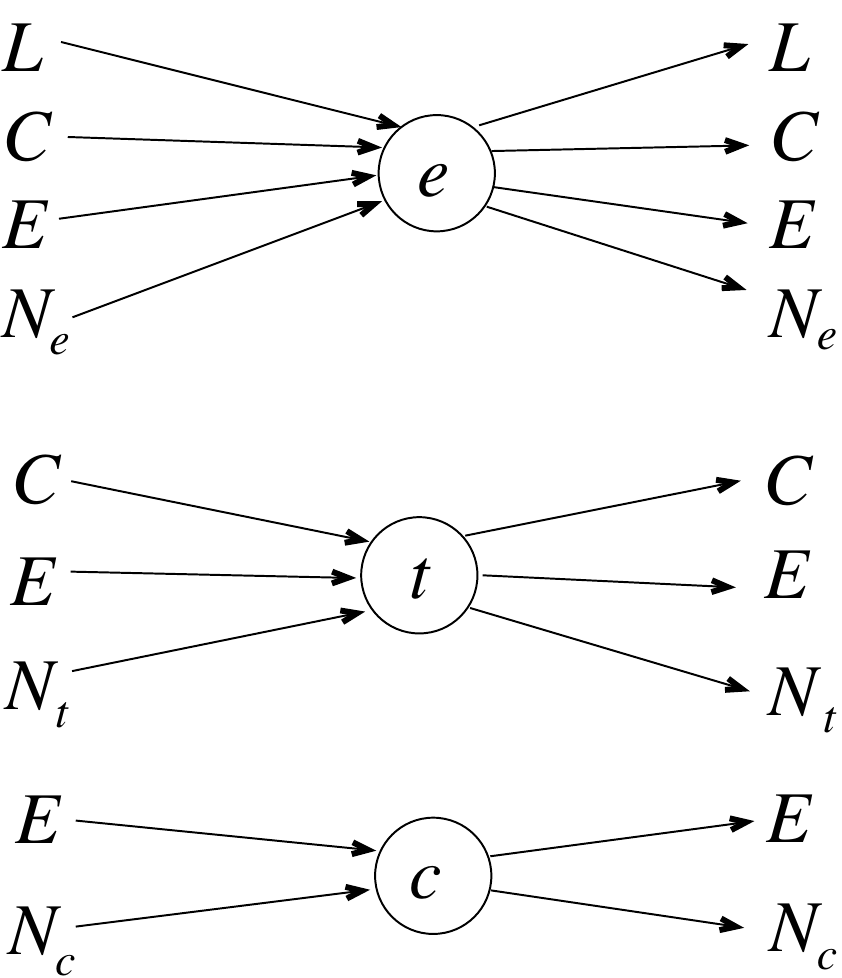}
\caption{\small Allowed semantic transitions through node types, by kinds of arrow. Not all of
the links are freely joinable, however, so there are restrictions on allowed transitions.\label{sst3}}
\end{center}
\end{figure}

\subsection{Inference rules and symmetries}

In an unconstrained graph there can be no rules for inference, because rules require
some regularity and functional predictability. In Semantic Spacetime, however, one
has four broad kinds of relation and three kinds of node or entity.

If we seek a strongly constrained deterministic logic, as in ontology approaches to graphs,
the result will be either simplistic or intractable.
Inference about common properties are straightforward, though remain speculative.
\begin{itemize}

\item If a collection of nodes $n_i$ are contained by a node $n_C$ which expresses property $P$,
then one might infer that the contained nodes {\em might also have this property}.

\item If a node has a property $P$ and is declared to be similar/near another node, then  
 the similar node {\em might also have this property}.

\end{itemize}
This possible inferential reasoning goes back to the discovery by Alva
Couch in join work\cite{stories,inferences}, which was subsequently
deepened in the development of semantic spacetime.  Unlike logical
ontological schemas, this kind of reasoning is simpler but inexact.
With ontological first order logic, results are either precise or non-existent.
In practice, most relations on data that are not carefully designed will fail to yield
any result due to the over-constrained nature of first order logic.

A second kind of symmetry concerns patterns of inference in the
linkage of nodes. Duplicate nodes may arise in a graph either by
accident or by a deliberate encoding of redundancy (degeneracy). If a
collection of nodes each possesses the same incoming and outgoing
links of a given ST-type, then we can infer that they are functionally
equivalent with respect to that process. This implies that they can be
treated as a single node (which we refer to as a supernode) for those
intents and purposes, though perhaps not all. While the nodes might have
the same links for, say, causal trajectory (`leads to'), they might have
different properties. This suggests another possible set of inferences
or warnings to flag: why is there a partial but not complete equivalence?
Is it intentional or accidental. Is it, in fact, an error?

\subsection{Absorption by blind alley and by statistical aggregation: scales and degrees of freedom}

How do we draw a ring around a region of interest or influence in a graph?
The role of an agent boundary in determining what is a value or operation in 
a given context is important to many of the concepts as we scale up or down
a hierarchy of meaning. Concepts emerge by recombination of atomic or genetic concepts
and attributes, suggesting that $C,E$ spatial types have semantic limits.
$L$ may have frequently have practical limits due to the ephemeral nature of
interactions and processes in general, but there is no obvious upper limit to the
extent of time.

The structure implies by the SST $\gamma(3,4)$ model is of a graph
composed of snaking causal sequences of events or ST-type 1
(`leads-to') connected nodes, where each node has `contains' and `express
property' nodes in orbit around them, expressing their interior
attributes.  These may themselves be in orbit around nodes that
contain or express them.  A few `nearness' links provide shortcuts
(wormholes) between nodes that are marked as being similar for reasons
outside the scope of explicit causal process knowledge.

One can say that events effectively act as bipartite bridging nodes between
invariant things and concepts, leading to a temporary connection in the sense that
an event has a finite contextual validity or lifetime even if the historical node remains
in the graph.

Aborbing regions arise whenever there are source nodes or dead end sinks in a 
directed graph. Once we descend into the details of a node, by {\sc contains} or {\sc express},
we find their invariant attributes. They live in an interior subspace orthogonal
to the timeline of events. The timelike ST-type 1
(`leads-to') are analogous to the Hamiltonian evolution in symplectic systems of physics.
This orthogonal subspace is analogous to the hidden dimensions of a Kaluza-Klein
or string theory, a field space of forces or `interaction semantics'.
The entire directional links for containment and expression point to ultimately aborbing
regions, as they are spacelike. 

There is an erasure of information in two ways here:
\begin{itemize}
\item Flows that end up pooling at sink nodes are aggregated with a loss of distinguishable
history or identity.

\item Processes that drive several nodes towards an appointed successor
merge separate dynamical flows rendering them indistinguishable.
Conversely, a node that splits a flow to several successors with a weighted
distribution is either sharing (dividing) or amplifying (multiplying) the
flow along the links.
\end{itemize}
Clearly driving independent flows through a common node erases
information and semantics, coarse graining the result. If one cares to
measure the incoming distribution against the outgoing, with an entropy
function, there is a change of entropy--potentially an increase or a reduction
depending on how one chooses to define the semantics of the result.
One does not avoid process semantic issues merely by adopting a conventional
narrative.

The significance of absorbing regions, sources and sinks, for a finite
system is clear semantically. Dynamically,
however, there are different ways to interpret
the processes.  

There are certain elements that are, in a sense,
atomic or irreducible by virtue of being a logical end to a process.
States with both incoming and outgoing arrows are transitory but
reducible ultimately to the sets of nodes either leading to it or
emerging from it. In differential calculus one refers to these
processes as retarded and advanced solutions. 

Equilibrium solutions are
also possible by selecting from both sets of arrow, or removing the
arrow direction altogether.  These are known as Feynman boundary
conditions in physics. The latter is interesting because it introduces
a second type of absorption for a process: a statistical absorption,
which is associated with {\em entropy}. Any bulk state of a system,
which converges statistically by some separation of scales into an
average {\em probability distribution} of values, may be called
statistically stable\cite{grimmett1}. For example, an ideal gas at
finite temperature has a statistically stable (or maximum entropy)
distribution of particle velocities, which it makes no sense to count
as distinguishable phenomena.  This, at the semantic level of the bulk
gas, the thermal state is absorbing---and represents another kind of `zero':
placing a small object into a large (fabled infinite) reservoir at temperature $T_0$ leads to
the process:
\beq
\text{Equilibrium } T_\text{object} = T_0.
\eeq
In other words, semantically, equilibration is another kind of zero operation.

\section{Matrix representations of graphs}\label{matrices}

A graph has an associated matrix representation for mapping
the topology onto a linear map acting on rings or fields. The matrix
representation is a crucialmap between arithemtic algeabraic rules and the
realization of processes as graphical structures.

The {\em adjacency matrix} $A$ and its transpose $A^T$ are square
matrices, whose rows and columns are the node labels, and whose
non-negative elements represent quantitative link weights.  The
numerical values are typically set to 1 in elementary texts on graph
theory, but they can have relative weights as well as semantic labels.
These may be chosen as non-negative real numbers. An {\em undirected}
graph (with arrows in both directions) is symmetrical about the
leading diagonal, i.e. $A = A^T$. A directed graph is asymmetrical and
its source and sink nodes, which are the starts and ends of paths,
lead to zero rows and columns, which consequently lead to zero
eigenvalues of the matrix. This proves to be important in a number of
ways, as it implies the matrix is non-invertible, preventing
predictive process-path reversals.  A complementary {\em line} or
`join' graph matrix is the complement of the adjacency matrix, where
rows and columns are represented by the links joining nodes. Both of
these may be found from the incidence matrix.

The {\em incidence matrix} and its complement describe the
  emission and absorption of link lines from and to nodes. Since they are local
to a single node, they may be associated with the promised intent of the nodes concerned. The
  standard conventions in most texts are for undirected graphs and are
  unhelpful here. Directed and labelled graphs with self-referential
  loops require a separation of the incident matrix into two parts.
  These are analogous to the offer and acceptance promises in Promise
  Theory, and the matrix elements factor from a Hadamard product form into 
two complementary matrices that are effectively the square roots of the
adjacency matrix: $I^{(+)}$ and $I^{(-)}$.
\beq
\hat I^{(+)}\hat I^{(-)} = \hat A + \hat C
\eeq
for some diagonal matrix $C$ belonging to the Cartan subalgebra of the flow.
When there are no auto-referential (pumping) self-loops, $C$ is proportional to
the identity matrix. The incidence matrices $I^{(+)}$ and $I^{(-)}$
correspond to the Promise Theoretic {\em offer} and {\em acceptance} promise rates\cite{promisebook}:
\beq
\hat I^{(+)} &=& \bordermatrix{
~ &  L & C & E & N_e & N_t & N_c  \cr
e &  1 & 1 & 1 &  1  &  0  &  0   \cr 
t &  0 & 1 & 1 &  0  &  1  &  0 \cr
c &  0 & 0 & 1 &  0  &  0  &  1  \cr
}\\
{\hat I^{(-)}} &=& \bordermatrix{
~    & e & t & c \cr
L    & 1 & 0 & 0 \cr
C    & 1 & 1 & 0 \cr
E    & 1 & 0 & 1 \cr
N_e  & 1 & 0 & 0 \cr
N_t  & 0 & 1 & 0 \cr
N_c  & 0 & 0 & 1 \cr
}
\eeq
Notice that these are not simple transverses of one another, as they would be
in a simple undirected graph, since a thing cannot be expressed (a forbidden state transition).
We distinguish nearness links for each of the three types $e,t,c$ for convenience.

Rates of change within the processes can be represented as derivative.
The `dynamical' graph derivative for node values is defined $\grad_i v_j = v_i - v_j$.
This corresponds to the usual Newtonian derivative $\partial_x v(x)$ for a function 
which is distributed over graph nodes $\vec v(N)$. 
There is a second notion of rate of change for a graph: 
because links define both direction and value between each pair
of nodes, they also behave as a vector field, which has a gradient role of its own.
The matric of links mapping to links forms a dual `line graph' representation
in which there are rows and columns for every independent link, no matter the
nodes they connect.

\subsection{Matrices for $\gamma(3,4)$ skeleton}

The $\gamma(3,4)$ skeleton graph can be represented, without explicit nodes only type
names, as a set of matrices characterizing the graph. 
Certain rules about process semantics mean that transitions between certain node types
are limited to specific kinds of arrow:
\beq
\hat A = A(n_i \mapsto n_j) ~~~~~~ &=&  ~~~~~~
\bordermatrix{
~ & e                     & t                & c \cr
e & \pm L, \pm C,\pm E, N_e  & +C               & + E \cr
t &  -C                   & \pm C,N_t        & + E \cr
c & - E                 & -E               & \pm E, N_c\cr
}
\eeq
This can be decomposed into a number of generators with antisymmetric (or anti-Hermitian)
signatures:
\beq
\hat A_L&=&
\bordermatrix{
~ & e                     & t                & c \cr
e & \pm 1 & 0 & 0 \cr
t &  0 & 0 & 0 \cr
c &  0 & 0 & 0\cr
}\\
\hat A_C&=&
\bordermatrix{
~ & e                     & t                & c \cr
e &  \pm 1 & 1 & 0 \cr
t &  -1 & \pm 1 & 0 \cr
c &  0 & 0 & 0\cr
}\\
\hat A_E&=&
\bordermatrix{
~ & e                     & t                & c \cr
e &  \pm  1 & 0 &- 1 \cr
t &     0 & 0 & +1 \cr
c & - 1 & -1 & \pm 1\cr
}\\
\hat A_N&=&
\bordermatrix{
~ & e                     & t                & c \cr
e & 1_e & 0 & 0 \cr
t & 0 & 1_t & 0 \cr
c & 0 & 0 & 1_c\cr
}
\eeq
Conversely, arrows in a trajectory can only be joined by certain types of node
as a path join matrix $J$:
\beq
J_{\gamma(3,4)}\left(\promise{a} n_i \promise{a'}\right)  ~~~~~~ =  ~~~~~~
\bordermatrix{
~ & L         & C       & E     & N  \cr
L & e         & e       & e     & e  \cr
C & e         & e,t     & e,t   & e,t \cr
E & e         & e,t     & e,t,c & e,t,c \cr
N & e         & e,t     & e,t,c & e,t,c \cr
}
\eeq

\begin{itemize}
\item A process $L$ must terminate on a final event or never.
\item A property attribute process $E$ must terminate on an atomic concept (property).
\item A containment process $C$ must terminate on an atomic thing (component).
\item A metric similarity process $N$ need not terminate within the scope of connected nodes.
\end{itemize}

The key point for directed graphs, representing finite processes of different types, 
is the existence of isolated states that absorb and emit transitions of the graph.
In a ring or field this is not the usual case. There one has `translational invariance'
or group transformations that are essentially unlimited, for both addition and multiplication,
with the important exception of the zero element, which is the one absorbing state under 
multiplication.

\subsection{Stable regions of a graph, information, emission and absorption}

Particularly for directed graphs, but also for some undirected ones,
an important feature within a graph is the number of nodes at which
the link flow converges (when absorbed by a node) or diverges (when
emitted by a node), i.e. for in- and out-degrees greater than 1.
These are confluence and branching points in the arrow vector field, 
which are literal and metaphorical singularities. As remarked in \cite{jan60}, 
emission and absorption by nodes is associated with zero operations.

There are certain operators which transmute interior state to exterior
movement in the graph.  These are related to ladder operators known in
the context of differential equations, and they are further connected
to absorbing states of the graph.  Convergent semantics with
idempotence of an operation one some end state comes about from ending
up in a cycle that has the identity element as its final state.
This is used to good effect in enforcing policy choices\cite{jan60}:
\beq
\hat O \vec v   &=& \vec v_0\nonumber\\
\hat O \vec v_0 &=& \vec v_0
\eeq

Consider figure \ref{confluence}:
\begin{figure}[ht]
\begin{center}
\includegraphics[width=4cm]{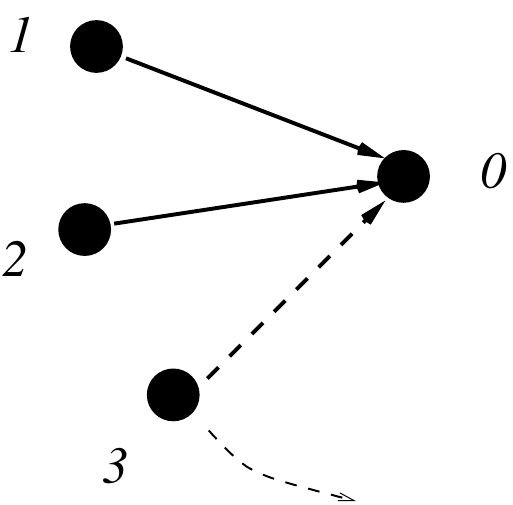}
\caption{\small A convergence of flows at a point is an absorbing region
of the graph, associated with a singularity.\label{confluence}}
\end{center}
\end{figure}
We can write this algebraically as:
\beq
\text{Arrow } n_1 &=& n_0\nonumber\\
\text{Arrow } n_2 &=& n_0\nonumber\\
\text{Arrow } n_3 &=& n_0,
\eeq
which is clearly isomorphic to
\beq
0 \cdot 1 &=& 0 \nonumber\\
0 \cdot 2 &=& 0 \nonumber\\
0 \cdot 3 &=& 0.\label{zero}
\eeq
in other words, by virtue of ending up at the same state `0',
the arrows lead the process into a location which does not remember
the route by which is arrived there.
If there are no arrows flowing away from node 0 then the node is
completely absorbing (a kind of black hole for the process), and it is
called a {\em sink} node. The adjoint, in which arrows are reversed would be a node that
emits arrows from nowhere, called a {\em source}.
Even with arrows both incoming and outgoing, any such multi-line convergence
is singular, and indeed this is represented in the adjacency matrix of the graph
by the existence of zero eigenvalues--and, indeed, the implications for invertability
of the zero operations. Inverse of zero is a topic that several authors have discussed.

A hub operation of this kind turns a distribution into a single value
and vice versa.  Clearly, one can turn $n$ values into one by
averaging or some selection process, but turning one value into $n$
requires more information. One can duplicate, triplicate, etc values
to send out identically to multiple redundant destinations. Then the
distribution of values is unchanged. Alternatively, one can use a new
source of information in different directions to determine the result
in different directions. One possibility is to use a differentiated
distribution of link weights. However one chooses to inject
information, it has to come from outside the starting node.

With multiple states distinguished semantically, rather than representing
different locations on a neutral number line, we now have more responsibility
to define the degrees of freedom carefully. In physics, one uses entropy concepts
to describe and measure the extent and homogeneity of a distribution over states.
The Shannon entropy\cite{shannon1,cover1} is defined over some distribution of states
partitioned into $N$ choices, and measure each with an alphabet of states $1 \ldots C$.
\beq
p_i &=& x_i / \sum_i x_i\\
S &=& - \sum_{i=1}^N p_i\log_C p_i.
\eeq
This is maximal $S=\log_C N$, when $p_i = 1/C,~ \forall i$, and minimal $S = 0$ when $p_i = 1$ for
some choice. If we increase the resolution or alphabet of the distribution $N \rightarrow \infty$,
then the entropy gets larger. If we coarse grain by making $C\rightarrow \infty$ then the entropy
approaches zero and then ceases to be defined for normal arithmetic
rules\cite{bergstra2025expressingentropycrossentropyexpansions}. The ambiguity lies in how
we count the implicit dimension of the state space (see figure \ref{sst1} and the discussion below).
The issue here is that a complicated process has more attributes or `degrees of freedom' than a simplistic view
of the number line.

In thermodynamic statistical physics, the Boltzmann entropy mimics the
Clausius entropy as a measure of the energy in a system, which by
virtue of being distributed indistinguishably around the system, has
lost its capacity to do work. This is because statistical absorption
is just as powerful as absorption by a single state or node of a
process. In either case, the ability to make distinctions over a
prescribed scale is the relevant issue.  No measure of information can
be conserved by processes that converges or diverges into a change of
degrees of freedom, on any scale, and hence cannot be restored without an injection
new boundary information for the inverse operation at or above the scale of
aggregation.

\subsection{Sources and sinks}

A state is absorbing in a semantic graph because this reflects the
interpretation we {\em intend} for it.  In other words, it's no accident
that we end up with absorption. The same is true in arithmetic if one
is careful, but there are cases where one is led to seek answers in
ways that bump into problems concerning the incompleteness of
definition.

If we think if the values in the field as nodes in a graph, with
arbitrary many nodes, then operations that take us from one value to
another may be represented as links with particular arrows that
represent the operational semantics.

Graph transformations are used widely in machine learning diffusion models for image reconstruction
and enhancement. They are also of interest here in a process knowledge representation for tracing the contextual
relevance of the map, which I'll return to below. 

Consider nodes 0,1,2 in the confluent junction in figure \ref{confluence}. The partial adjacency matrix is:
\beq
\hat A =
\left(
\begin{array}{ccc}
 0 & 0 & 0 \\
 1 & 0 & 0 \\
 1 & 0 & 0 \\
\end{array}
\right)
\eeq
From this, the transposed adjacency acts as a forward stepping operator over the vector landscape of internal node values,
and the untransposed matrix is a backwards stepping operator when acting on internal graph node values, represented
as a vector $\vec v^T = (v_1,v_2,\ldots)$ :
\beq
\hat \hat F =
\left(
\begin{array}{ccc}
 0 & f_1 & f_2 \\
 0 & 0 & 0 \\
 0 & 0 & 0 \\
\end{array}
\right)
~~~~~~~~
\hat \hat B =
\left(
\begin{array}{ccc}
 0 & 0 & 0 \\
 b_1 & 0 & 0 \\
 b_1 & 0 & 0 \\
\end{array}
\right),
\eeq
so that
\beq
\hat F \vec v =
\left(
\begin{array}{ccc}
 0 & f_1 & f_2 \\
 0 & 0 & 0 \\
 0 & 0 & 0 \\
\end{array}
\right)
\left(
\begin{array}{c}
 x_1 \\
 x_2 \\
 x_3 \\
\end{array}
\right)
=
\left(
\begin{array}{c}
 (f_1x_1+f_2x_2) \\
 0 \\
 0 \\
\end{array}
\right).
\eeq
We note that $\hat B$ and $\hat F$ are not inverses of one another, since they both contain
zero eigenvalues, i.e. have determinants of zero.
Operating on the graph's internal state with this stepping operator, we see that the values
from nodes 1 and 2 are shunted onto node 0, with the weighting determined by the adjacency
matrix link weights. The original value at node 0 falls off the end of the absorbing node into the void
and is unrecoverable. If we now try to reverse this in order to restore the original information,
we see that the absorbing node wipes out the memory of the system rendering an inverse impossible without
an input of new information:
\beq
\hat B\hat F \vec v =
\left(
\begin{array}{ccc}
 0 & 0 & 0 \\
 b_1 & 0 & 0 \\
 b_2 & 0 & 0 \\
\end{array}
\right)
\left(
\begin{array}{c}
 f_1x_1+f_2x_2 \\
 0 \\
 0 \\
\end{array}
\right)
=
\left(
\begin{array}{c}
 0 \\
 b_1(f_1x_1+f_2x_2) \\
 b_2(f_1x_1+f_2x_2) \\
\end{array}
\right).
\eeq
If we select the values for $\vec f$ and $\vec b$ appropriately, we can partially restore the original state, but
not without specific knowledge of the original configuration and the ability to inject appropriate values
into other nodes.
The absorbing node $x_3$'s value is lost forever unless we insert a value by hand (as a matter of policy).
and the initially unused source value $x_1$ has been injected.
The values of $b$ would typically involve division by the dimension or node degree $k_in$ of the absorbing junction node $0$.
\beq
(f_1x_1+f_2x_2) \rightarrow 1, b/f \rightarrow \frac{1}{2}.
\eeq
Why is the reverse operation not equal to the inverse matrix?
This can be traced to the zero eigenvalues (and zero columns in the $\hat F$ operator). The computation of
a direct inverse would require a division by zero, which could yield any value, from eqn (\ref{zero}).

For a $3\times 3$ matrix we can write the inverse explicitly for arbitrary real numbers $a,b,c,d,e,f,g,h,i,j$:
\beq
M = \left(
\begin{array}{ccc}
a & b & c\\
d & e & f\\
h & i & j\\
\end{array}
\right)
\eeq
The transposed cofactor matrix (adjugate) forms the the linear combinations 
which can be reverse engineered to yield cancellations
or determinants, giving a formula:
\beq
M^{-1} = \frac{1}{\text{det}(M)}\left(
\begin{array}{ccc}
(ej-if) & -(bj-ic) & (bf-ec)\\
-(dj-hf) & (aj-hc) & (af-dc)\\
(di-he) & -(ai-hb) & (ae-db)\\
\end{array}
\right)
\eeq
where $\text{det}(M)=a(ej-if)-b(dj-hf)+c(di-he)$.
The difficulty arises in the scaling of the values to renormalize the diagonal to $\vec 1$:
\beq
M\,M^{T} = M^{T}\,M = I = \vec 1.
\eeq
For a sparse graph, most of the values $a,b,\ldots$ are zero. The ability to
evaluate the expression is then unclear, since the rules for rings and fields
do not admit division by zero. Here there are so many zeroes in multiplication
that one has to see them as strings of operations rather than mere numbers.

The renormalization of the inverse is related to an overall factor of the determinant of the matrix.
Since the determinant is the product of the eigenvalues, it is zero if there if a
zero eigenvalue, which corresponds to a zero row or column. For
a directed graph, this corresponds to a node which goes nowhere  i.e. an aborbing state.

The effect of $\hat B\hat F$ is to eliminate the degree of freedom at
$x_0$ and to coarse grain the values of the others so that they become
equal. The only way to restore the original distribution is to encode
the original values by a memory operation $\hat M(\vec x,\hat F)
\rightarrow\hat B$. Indeed, this is schematically how diffusion models
of machine learning restore images: by training an inverse process to
capture the process memory of destroying the image through insertion
of noise.  The presence of $0^{-1}$ in the determinant or inverse
indicates the need to remember past history, a snapshot of the past to
`roll back' to, as pointed out in \cite{jan60}.

We can now associate the semantics of stepping and state exposure with the semantic spacetime properties.

The same inverse notions apply to the semantic graphs.  If we try to
shift context up or down a branching graph, in the information
hierarchy, crucial context may be lost. The relevance of the path is
reduced by the dimension of the possible alternative pathways.  This
loss is happening inhomogeneously all over the graph where a process
in ongoing. While this might initially seem harmless, making sense of
the result requires a continuous input of new information to keep the
inverse on course--but, on course to where? Deciding this selection
requires an intentional act, i.e. an intentional insertion of policy
information at each stage. In diffusion models, this is provided by
prompt information from a user.

As with logics, if one attempts to remember complete and precise
information, then one would only be able to generate results that were
explicitly input. No recombinative mixing or `lateral thinking' could
enter the process to create something new (however derivative).

In earlier work, I proposed a T-rank algorithm for maintaining a more
stable entropy distribution over a $\vec v$ by pumping graph emission
self-referentially to counter absorption. While this works, it also
emphasizes that an ad hoc input is needed to obtain a stable answer,
and that this ad hoc prescription basically determines the outcome
with possibly only a shadow of the original constraints intact.

\subsection{Frobenious-Perron eigenvector theorem}

The eigenvector equation
\beq
M \vec v = \lambda \vec v,\label{eigen}
\eeq
for some matrix $M$ and vector $\vec v$ has solutions called eigenvectors, which are
intrinsic (eigen) properties of the matrix in some sense. 
The equation can be applied to graph matrices by taking the adjacency matrix $M\mapsto A$,
which is non-negative, over the nodes of a graph $\Gamma(N,V)\mapsto (\vec v,A)$. This technique
is widely used in social network analysis to perform so-called importance ranking of nodes
in the graph\cite{graphpaper}.

The Frobenious-Perron theorem for non-negative matrices states that the largest or 
principal eigenvector of any such graph will be entirely positive. More significantly,
the semantics of this vector attach to the eigenvalue equation:
Iterating equation (\ref{eigen}) represents a recursive propagation of node values over links, weighted by the link 
values, which implies that multiplying any non-zero vector repeatedly by $A$ will converge 
$\vec v$ towards the principal eigenvector.

The purely positive addition of purely positive values must yield the highest eigenvalue.
An undirected graph is in flow equilibrium, so there is no net direction to the movement
of values over the links. The equilibrium distribution of $\vec v$ in the principal eigenvector
thus represents the reservoir water level at each of the nodes (flow capacitance) at equilibrium.
In social networks, this is equated with social capital or `importance ranking' (an important person
is someone with many important friends).

For an undirected graph, where node connections propagate without opposition, all the value
flows along directed paths to sink nodes, where it pools.
The normal algorithm for computing the eigenvector by operating many times with $A$ onto a vector of ones $\vec 1$
fails in this case to give the right answer, essentially due to the ambiguity of division by zero.
When there are zero rows (absorbing nodes) the effect of this computation is simply
zero. A more careful analysis shows that the vector components for absorbing nodes should
be non-zero as this is where all the flow piles up, but the multiplication by zero overwhelms the 
normalization of the vector by $\lambda$. which may be zero itself for the absorbing nodes yielding
$0/0 \mapsto 1$.

\section{Graph semantics of operational arithmetic computation}

We can illustrate some of the semantic choices in basic arithmetic
operations using graphs as the domain `space' to model what we mean by
the operations. The standard meanings are so ingrained in us from an
early age that it might seem strange to question them, yet doing so is
quite instructive.  The purpose of doing so is not to necessarily
change conventions or solve some inconsistencies, but to point out how
a few common themes trace back to interpretational ambiguities that we
take for granted in mundane arithmetic. These become crucial to our
understanding when we adopt semantically rich knowledge representations,
such as in artificial reasoning.

Without adopting the standard jargon of rings and fields, arithmetic
concerns the rules for counting and measuring amounts of `stuff'. It
uses formal quantities called $x$ to model `amounts' and handles
operations of augmenting and combining (+), depleting (-), duplicating
($\cdot$) and sharing ($/$) operations. These operations should work
in both quantitative and qualitative interpretations.  The
abstractions either expose or conceal information however.

In the algebra of rings and fields, the status of numbers as positions
or as shifts is blurred into a single set theoretic domain along the
number line.  Apart from this conventional interpretation, the
question of semantics remains ambiguous for graphical representations
of operations.  Should numbers be thought of as value locations or as
part of transformational operations?  Should we treat numbers as
events, things, or concepts? And perhaps more pertinently, do we
muddle these interpretations carelessly in dealing with numbers? This
question is particularly interesting in the service of Quantum Theory,
where operator algebras and numbers co-mingle extensively.

As an illustration of this, we can begin by looking at graph
representations of basic arithmetic reasoning and ask what are natural
interpretations for addition, subtraction, multiplication, and
division where the domain of mapping is no longer simply a single copy
of the number line.  Although this feels slightly self-indulgent, it's
illustrative of the fundamental issues in attributing semantics to
formal processes and therefore underpins everything else in a form
which is familiar to all readers.  Let's consider some examples, which
might not be exhaustive.

The first question for semantics is to ask how we actually mean to represent a number to be
added, subtracted, multiplied or divided. There is no unique answer
to this question, but we can naively imagine people counting pebbles or abacus beads.
The standard rules for calculating are guided by a principle of closure around
a set of numbers: when we combine numbers, the result should be a number of the same `type'.
\beq
+ : \R1 \times \R1 \mapsto \R1.\label{rrr}
\eeq
This is clear enough in the limited realm of mathematics, because the semantics of the 
ordered number line $\R1$ have already been defined  (see figure \ref{translate}).
The number line is a key visualization of the process that works well for a geometrical
interpretation of addition, but less well for a bulk interpretation.
In our standard definition of division, for example, which is based on the primacy of the
number line, there are `design issues' to consider: if we wish our answer to 
a division operation to remain (mapped back into) the
scope of the number line, then we can no longer both map an agent of
size $x$ back to an agent of size $x$ {\em and} split the original
inventory amount into $a$ parts for all values without expanding the 
concept of numbers to include fractional amounts. Once one introduces
new numbers, negative numbers and so forth, the consistency of the whole is jeopardized
unless one can close the operations convincingly. This had resulted in the
invention or discovery of symbols representing $i = \sqrt -1$, and more recently $\perp$ or $\Phi$ for concepts
including $1/0$\cite{bergstra2021divisionzerocommonmeadows,janderson1}.

\begin{figure}[ht]
\begin{center}
\includegraphics[width=11cm]{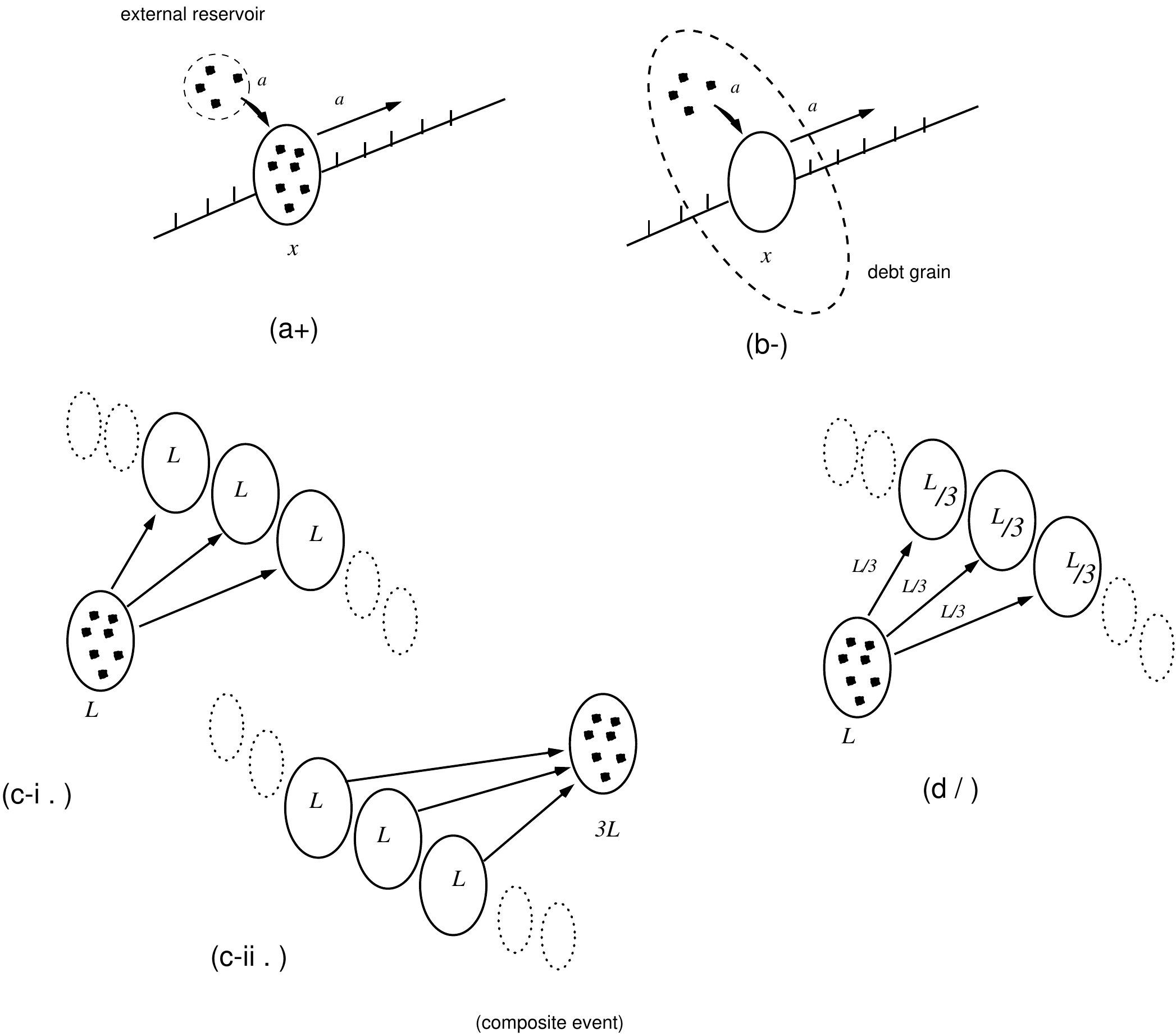}
\caption{\small Some semantic interpretations of addition,
  subtraction, multiplication, and division in graphical form. A graph
  has more degrees of freedom than the number line automorphisms of
  rings and fields.  Notice that multiplication has at least two possible
  interpretations: as an outgoing amplification of nodes (with an
  expansive dimensional meaning) or as an additive aggregation from
  nodes (mapping onto a single value more like the conventional
  ring/field interpretation).\label{arithmetic}}
\end{center}
\end{figure}

\subsection{Arithemetic}

Let's consider the arithmetic operations on the number line as an
example. These operations define operations on rational numbers in
graphical terms (see figure \ref{arithmetic}), but run into
difficulties with irrational numbers.  In order to cover irrational
numbers (which some mathematicians deny the existence
of) one has to introduce infinities and limits on the interior of
agents. These are fascinating issues that can only be mentioned in
passing here (some discussions can be found in
\cite{janderson1,janderson2,bergstra2,bergstra3,bergstra4,bergstra5}).

Is there such a thing as a pure number?  Arithmetic algebra defines
unary $1 \mapsto -1$ and binary $1+1 \mapsto 2$ operators, in which
the values are often pictured as locations on the real number line
$\R1^1$.  Euclidean geometry builds on this idea to coordinatize
$\R1^n$ and make the explicit connection between value and location in
a spatial construct. A graph is an analogous structure to a Euclidean
vector space

The algebraic structures of geometry and arithmetic are so ingrained
in daily norms that we seldom stop to confront the details of why
these structures work. Representing these processes as graphs is an
interesting exercise in self-consistent representation, as algebraic
graph treatments involve matrix arithmetic, which is a generalization
of ordinary arithmetic for partially coherent parallel processes.
The stories or explanations we tell about these operations sometimes
deviate from the actual rules and results provided by rings and
fields.  We can try to use graphical representations to elucidate the
intended meanings.

\begin{figure}[ht]
\begin{center}
\includegraphics[width=4.5cm]{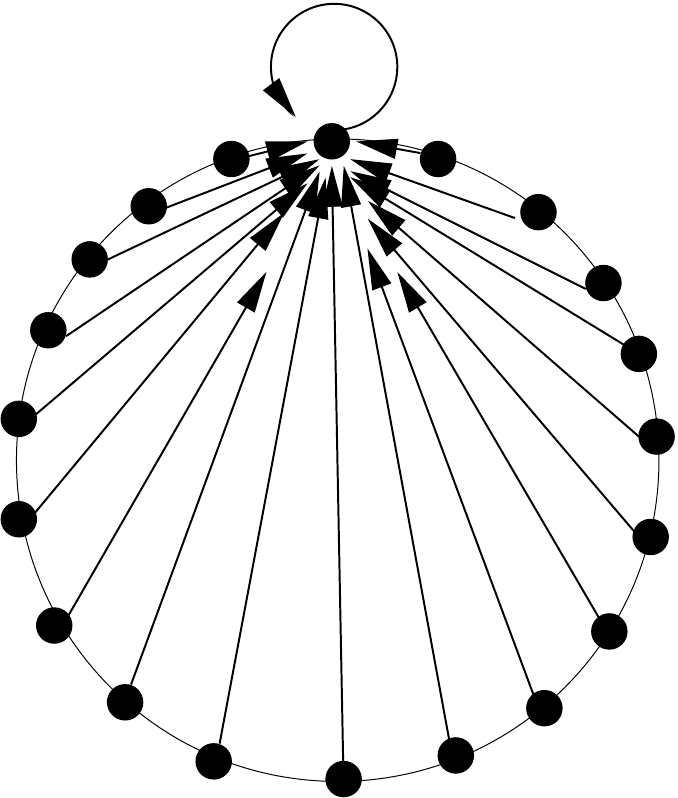}
\caption{A compact representation of a zero multiplication interpreted graphically on a circular
topology with $L$ rather than $\infty$ nodes. When inverting this, the number of possible
shares is represented by the number of arrows, like the  vector results in
eqns (\ref{remainder}) and (\ref{vector}). The common interpretation of $n/0 \rightarrow \infty$
could be associated with these arrows, so should the answer be $n/0 \rightarrow 1/L$ in a finite system?\label{sst1}}
\end{center}
\end{figure}

\subsection{Special arithmetic states 1 and 0}

Key to establishing the axioms and theorems of rings, fields, and groups are the special values
1 and 0. These binary values have attained legendary status in the digital age, but their significances
are more important than binary arithmetic. They are the two stable fixed points of
arithmetic:
\beq
1 \cdot 1 &=& 1\nonumber\\
n \cdot 1 &=& n\nonumber\\
n \cdot 0 &=& 0
\eeq
These are graphical processes (see figure \ref{arithmetic})
Here there is a special value 0 which is an absorbing state, i.e. arrows that enter do not leave
(figure \ref{sst1})).
\beq
0 \times 0 &\mapsto& 0\\
1 \times 0 &\mapsto& 0\\
2 \times 0 &\mapsto& 0\\
3 \times 0 &\mapsto& 0\\
4 \times 0 &\mapsto& 0\\
\ldots
L \times 0 &\mapsto& 0.
\eeq
In this interpretation, the proposal that $x / 0 \mapsto \infty$ is a
statement of the dimension or degeneracy of the inverse map. The
inverse map is not single values and thus a prescription for
interpreting it is needed. In machine learning diffusion models, for
example, this inverse is interpreted as a Markov process and a policy
decision is used to invert the map to get `something from nothing'.

More generally one can introduce a new value (something like the square root of minus 1) such that
$x/0 \equiv \perp$. But what kind of object is $\perp$?
In the finite graphical interpretation, one might imagine that the dimension of the result
would be $L$.

\subsection{Addition as a graphical process}

Translation is one dominant interpretation for additive arithmetic.
Consider the expression:
\beq
x' = x + a\label{xpa}
\eeq
where $x,a$ are what we intend to mean as `numbers' (counts or measures). 
This operation of addition has two common interpretations that are easily distinguished
in terms of agents. In order to use agents, we only need to assume that the set of agents is countable.
In this model A number may refer to:
\begin{enumerate}
\item A location of an agent on the number line $\R1$, which is a externalized total ordering
of agents according to their number proper identities. The meaning of addition is then to translate
or redefine the labels from coordinate position $x$ to position $x+a$. Relativity has implications
for the interpretation of these operations too.
\item The amount of interior holdings of the agent of some counter (e.g. money or energy etc).
The meaning of addition is then an augmentation of the holdings from $x$ to $x+a$. We say this is 
interior to the agent, because its position hasn't changed in our new interpretation. However,
one could also argue that this is the same as introducing additional exterior dimensions that
are private to our invariant meaning for location (as one does in Kaluza-Klein or string theories
of physics, for instance).
The question of
where $a$ new things come from, or how the semantics of $a$ (an increment) 
differ from the semantics of $x$ (a state) is typically brushed aside. 
\end{enumerate}
The difference between interior and exterior `locations', allowing us to distinguish exterior location
from interior holdings clearly has `boundary value' implications.

Indulging the forbearance of the reader for a moment, let's examine this in more detail, since the
issues are central to semantics of space and time and all graph based knowledge representations.
We take the first of these additive cases whimsically to mean a translation
along the number line, as in figure \ref{translate}:
\begin{figure}[ht]
\begin{center}
\includegraphics[width=12cm]{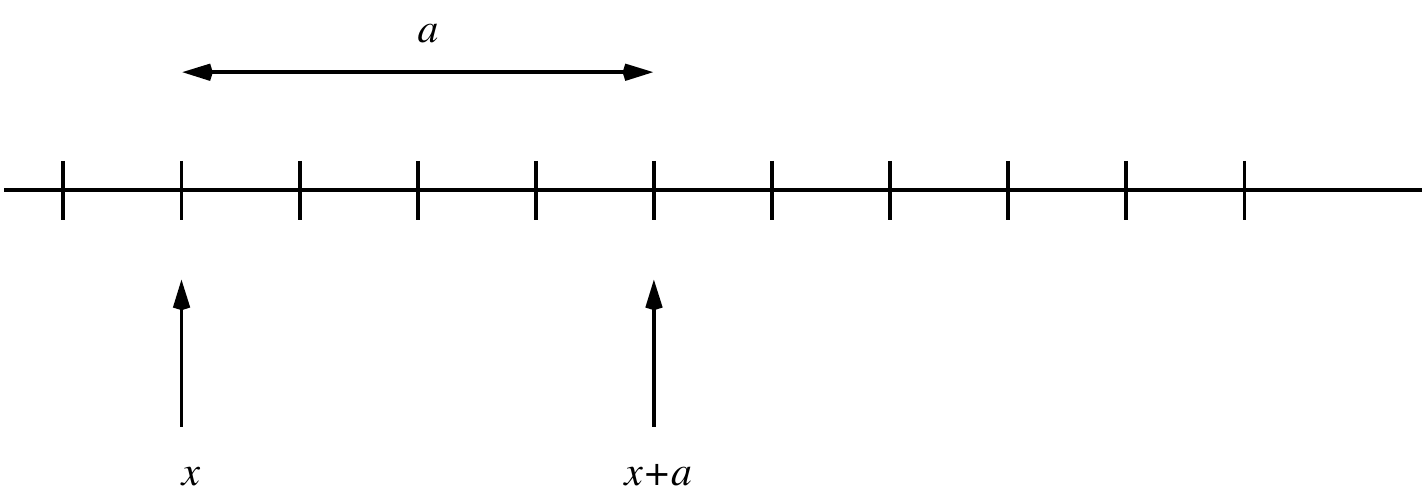}
\caption{\small Addition pictured as a geometric translation. This is a common image
in physics, where groups under addition are used to represent motion.\label{translate}}
\end{center}
\end{figure}
In most cases, we think of $x$ and $a$ as being representative of numbers belonging to the same set, not
to different copies of a set that looks like $\R1$, yet this is misleading if not inaccurate.
In both cases, the closure of the rational or real numbers under addition is a formal identification
of outcomes back onto the same space, so while we write (\ref{rrr}), we actually intend:
\beq
+ : \R1 \mapsto \R1 \times \R1 \mapsto \R1.
\eeq
The final map involves a loss of information. If we introduce an entropy of representation
For other operations one gets away with the same final mapping.

From a geometrical perspective we can think of $x$ as a position, and $a$ as being a argument to an operation
that shifts the state of our marker from $x$ to $x+a$. Implicit in this idea is, of course, the idea that
the successive values along $\R1^1$ are totally ordered so that the coordinatization is in one-to-one
correspondence with elements of some spacetime.
If we pedantically represent (\ref{xpa}) as an operation on $x$, then we could write:
\beq
\hat O_a\,x \mapsto x' ~~ :  ~~|x'| = |x+a|,
\eeq
i.e. an operation of type argument or magnitude $a$ on the initial position x $x$ leads to a new location, in which the location
$x'$ has a label whose value corresponds to $x+a$. Without clear semantics, this is only a tautology.
Indeed, it is only convention that would lead us to believe the result of operating on $x$ with $\hat O_a$ would
be a position and not some new thing.
This is not clear from (\ref{rrr}). 

Next, consider the interpretation as an interior amount.  
Aggregation is one kind of addition. We might call it `semantic addition' as opposed 
to addition as a translation along the number line. Now location
and translation are exchanged for inventory count and input amount,
which combine back to a new inventory amount. The question of where
the input comes from (a different place than the agent's own storage,
thus violating conservation) is sidestepped in the same way one
sidesteps the semantics of changes in thermodynamics: by inventing
fictitious infinite reservoirs that are approximately conserved and so on.
This is a familiar trick in handling arithmetical anomalous cases too, such
as with division.

To further illustrate
the points here, consider an alternative matrix representation of addition, which acts as a bridge between
the interior and exterior forms, since it uses an explicit extra dimension
to distinguish and represent the add and $a$ (see also \cite{jan60}).
\beq
\left( 
\begin{array}{c}
1\\
x'\\
\end{array}
\right)
=
\left( 
\begin{array}{cc}
1 & 0\\
a & 1\\
\end{array}
\right)
\left( 
\begin{array}{c}
1\\
x\\
\end{array}
\right)
\mapsto
\left( 
\begin{array}{c}
1\\
x + a\\
\end{array}
\right)\label{mxpa} 
\eeq 
In this form, $x$ and $a$ belong to different
`Euclidean' dimensions, or rows in the row space of the matrix.
Clearly the expression also applies even when $a = x$, so $2x = x+x$,
though the name $2x$ now assumes a new aspect of the algebra in linear
multiplication. An entire chain of interpretation is implicit in these
semantics.  In (\ref{mxpa}), the original meaning is embedded in the
new, so this is hardly progress. However, it serves to illustrate that
we can make representations as complicated as we like, with as many
dimensions as we see fit.  Matrices are of interest in this case,
because they offer the bridge between graphs and Euclidean vector
spaces.  We have intuitions about both, but these are easily muddled.

In a higher dimensional vector space $\R1^n$ or a graph $\Gamma(N,L)$,
we need to select a direction from the possible independent directions
available. Directional generators in $\R1^n$ can be made easily from
suitable matrices or tensors in a given coordinate system, assuming
homogeneous and isotropic spaces. For a graph it is generally more
complicated, because there is no consistent set of directions from
which to choose: each point points to a select set of neighbour
destinations, which is typically different at every node. At the same
time, the entire graph adjacency matrix can be thought of as a
stepping operator for propagating interior field values from node to
node\cite{graphpaper}.

In a graphical representation, the superficial similarities between
$\Gamma(N,L)$ and $\R1^n$ are clear, and are frequently exploited in
embeddings, e.g. for artificial intelligence feature representations
for artificial neural networks, in which one plays with interior and
exterior dimensional representations freely to enable independent
processes.

\subsection{Subtraction as a graphical process}

Subtraction semantics $x-a$ imply taking away part of a measurable
state. Once again, our interpretation will depend on the how we give
meaning to $x$ itself. As a translation, subtraction is the same as addition
with only a reversal of direction (assuming that the reverse direction
exists, which is assumed in vector spaces but is not generally true in
graphs or vector fields). This is a convenient group theoretic property,
because we need an inverse for every operation to complete a symmetry.

If, on the other hand, $x$ refers to the amount of inventory held by
an agent there is no opposite direction for filling states, unless one
take on the concept of debt.  This occurs in physics too because of
the attachment to a principle of conservation.  Whether the
conservation principles are actual physical truths about the universe
or simply self-consistent ways of counting slowly varying bulk
quantities is probably still unresolved. The changes in the way money
is allocated point to a similar issue. Money is not conserved but we
calculate with it as if it were.  This is the purpose of rings and
fields: to uphold a simple set of semantics that broadly enforce
conservation of amounts.

The addition and subtraction thus have similar semantics but with
opposite effect on the final states of an agent. A translation
forwards is similar to a translation backwards.  An addition of
inventory is similar to a removal of inventory, which the same
question about what become of the balance of the change. Apart from
the change of flow direction for $a$, subtraction is not substantially
different from addition until we reach amounts that involve
subtracting from zero.  This leads back to the question of
conservation and into what `afterlife' these amounts $a$ actually go to.  The
concept of `debt' is based on this conundrum, and the resolution
involves playing with coarse grains of time over which balance is
restored and ordering can be sacrificed for the greater good of
maintaining the conservation mythology. Subtraction then forces us to
embrace a wider world of dynamical behaviours, detailed balances, and
on going interactions with external reservoirs in order to give these
ideas meaning.  In doing so, it forces and explicit introduction of
averaging over time or space, repackaged by sleight of hand into
coarse grained or statistical conservation.

If we return to the translation interpretation of addition, pushing beyond the end
of the number line would mean losing the value altogether.  Topology
offers alternatives here. Either translations fall off the end of a
finite number line into a void, or one might identify the ends into a toroidal or
spherical topology (a circle in one dimension), in which case a
translation enters into the realm of modulo (`clock') arithmetic,
wrapping around with remainders. All calendar phenomena follow this
approach, with finite repetitions.  Multi-valuedness of counting
functions on the finite space is avoided by introducing more and more
dimensions to the clock (minutes,hours,days,years, etc) for the
parameters to spread into to answer the question of where the extra
information goes.  This is equivalent to adding new nodes to a graph
to prevent paths from coming to an end.

Finally, if one is unable to map all values back into a closed compact
set of dimensions as in a ring, then one may attribute the growth to
information loss by introducing a coarse-graining concept for garbage
collection to wipe away unrecoverable information as an undesirable.
Again, thermodynamics (which is an explicit agent model of energy
phenomena) has confronted the quantitative aspects of these issues
before.  Our goal here is to extend this to the semantic aspects too.

The semantics of conservation quickly point out an unavoidable link between
statistical (aggregate) information and elementary changes, which
leads to ideas of entropy.

In both cases, the algebra of rings gives a convenient answer to a
question, but by riding roughshod over specific semantics and
normalizing a response. The answer provided however is not necessarily
an an answer to the question we intended to ask. Theoreticians learn to
be cautious in the use of these conventions.

\subsection{Multiplication as a graphical process}

Multiplication addresses the duplication, triplication, and
$n$-ification of an amount. However, it also plays a role in the
meaning of repeated addition. One sees both interpretations
in play in mathematics: multiple addition has `translational' semantics, while
multiplication of directions has Cartesian `direct product' semantics.
Again, our understanding stems from integer
amounts, and generalizes conceptually to real numbers later. Once
defined for integers, ring and field algebras allow one to close
multiplication by non-integer values. (figure \ref{arithmetic})

Once again there are interior and exterior interpretations of
multiplication as an operation.  What aspect of $x$ do we wish to
multiply?  Referring to figure \ref{arithmetic}c, we could start with
an agent whose inventory is $x$ and wish to multiply the inventory
amount within the agent, or we could imagine a multiplication of the
agent itself, as a container, containing the equal amounts of
inventory.  No now $x$ is an agent with certain holdings, and $a$ is
an operational parameter with no direct connection to the number line.
It is not a distance. It's semantics are new.

As before, with addition, we conventionally choose that $x$
and $a$ and $x\cdot a$ all map to the same space $\R1$, and so (in
agent terms) we want to map each starting agent to $a$ copies that are
of the same type, and importantly contain the same amount of
inventory. This only works in practice for integer multiples, but we
can then postulate the concept of partial agents to complete the
calculational picture, in the same way that we use debt to delay the
realization of the amount in practice.

Once again, these issues are harder to avoid in physics, and have been confronted
in thermodynamics and statistical physics by Boltzmann and others. There are deep connections here to the 
notion of distinguishable and indistinguishable states and `particles'.

In ring algebra, we can think of multiplication as as successive addition. This is a result of
the final automorphism, mapping outcomes back onto a the original number line. While an external
observer can attest to the consistency of the answer in a counting scheme, several additions
are very different from the commonplace everyday semantics of multiplication (which involves manufacturing
new copies of a state) to make more things.

Conveniently, the association of inverse multiplication with division helps to cement this.
A similar association occurs in the Fundamental Theorem of Calculus, which proves that
indefinite (symbolic) integration is the inverse of differentiation and vice versa.
The semantics of differentiation as a gradient, and integration as a summation do not
obviously make this clear, yet there is a methodological precision (involving the
infinitesimal limits of intervals that allows the operations to be co-related.

What single aspect of the operation should be mapped back into $\R1$:
is it the amount received by each agent, the number of copies, or the
total inventory of all copies?

\subsection{Division}

Last but certainly not least, the semantics of division drive us more
forcibly into the realm of agent interpretations (figure
\ref{arithmetic}).  Division has no obvious translational
interpretation, but it can be related to the scaling and partitioning
of distances---dividing a journey into a number of `legs', or a procedure
into a number of subroutines.  This interpretation is
connected to the renormalization group in mathematical
physics\cite{scale1}.

A more common conception for division is simply to share a bulk amount
between a number of recipients, e.g. in a marketplace. In agent terms,
there is (or at least could be) a container or binding force to hold
the multiple separable parts in different `buckets'. In the absence of
further information, a sharing out of $x$ into $a$ parts implies that
the amounts are $a$-furcated into equal measures. There is no
particular reason why the measures should be equal, except for a sense
of simplicity and symmetry.  Prime numbers become an immediate source
of worry: what if the inventory is already atomic or indivisible, how
can we share amounts that do not divide exactly? Already we are faced
with the invention of real number semantics or the concept of
remainders.  What does it mean to share out the holdings of an agent
into equal parts of the same size, with or without remainders?
A remainder turns a division not into a single number, but a doublet
of (share per recipient,remainder):
\beq
5 / 3 \mapsto (1,2) \label{remainder}
\eeq
Matters of definition play an underestimatable role.
Division, like multiplication, does not yield a single unambiguous interpretation for its answer.
Are we counting the number of partitions, the total amount of stuff shared out, or the size
of an individual share? The answer could, after all, refer to either the number
of agents with equal shares or the amount within each agent.

The usual answer gives a value which is the average size of the original amount given to each
of the agents sharing it. Since this is equal, by argument, there is a unique value that can be
mapped back into the original system of numbers. It's meaning is lost, however.
If we share a proton between three agents, they would all get different quarks with
unequal properties. Only by throwing away information can the result be a single value
rather than a vector of the partitioning.

\begin{figure}[ht]
\begin{center}
\includegraphics[width=12cm]{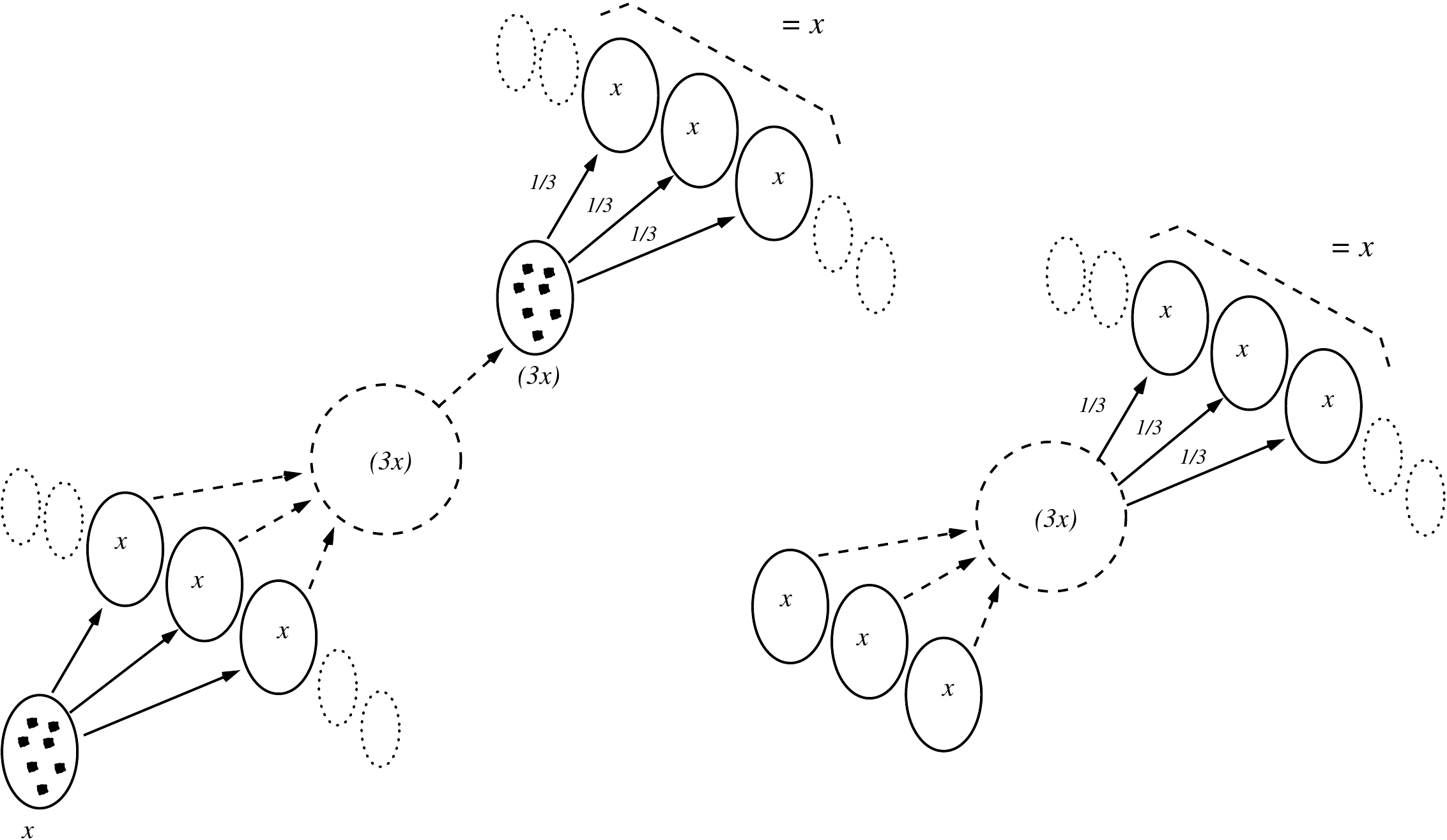}
\caption{\small Multiplication and division as inverse processes. Here we represent
$3x/3 = x$ as a process graph.\label{inversediv}}
\end{center}
\end{figure}

Division by $N$ is a partitioning of the a set into $N$ parts. This
makes obvious sense for all numbers except for numbers less than 1.
For example, division by 2 leads to two sets that are defined to be
equivalent and thus the return value is the result for one of them. We
do not write:
\beq
4 / 2 \not= (2,2) \label{vector}
\eeq
This could be interpreted in two ways: the 2 could mean each received a share of 2 or that there are two
copies (a `redundancy' or `degeneracy' of 2).
What about values smaller than 1? Dividing a set into half agents would mean that there are
\beq
\frac{~~x~~}{\2}
\eeq
implies that half agents would be twice as many
\beq
\frac{x}{ 0.5}  = ?
\eeq
The conventional answer of $2x$ here involves a renormalization of amounts. It is not so much
an answer as a redefinition of the scale used for the answer. Again the semantic legerdemain
leads to the convenience of an inverse operation to multiplication, at the expense of 
metaphysical caution.

\subsection{Division by zero semantics}

The more fraught issue of division by zero must also be handled by a
more careful convention.  It's possible interpretations are intimately caught up
in the processes it might represent---which should issue a warning to those who wish to
fully abstract a universal meaning for numbers.
Dividing a set into zero parts does not have
an obvious mundane meaning. It could mean that the result is nothing
as no-one receives any share of the initial state, or it could mean
everything since the initial amount remains undivided. Alternatively,
one could think of the limit $1/0 \rightarrow \infty$ as the
metaphysical effort or force needed to break through the barrier of
zero recipients, punching a hole through to the number line. That's an
interesting picture but with semantics that apply only to certain
process scenarios.

In another sense, multiplication and division share many similarities,
not only as adopted inverses for fields. Both involve nodes that can
be considered to absorb or redistribute values, with arrows
responsible for moving, aggregating, combining, distributing, and even
creating and deleting information based on the topology and the
semantics of the graph.  

Conservation of some counter (flow continuity) is one of the most common
criteria for defining processes, in order to have a consistent normalization
of quantities. This is pervasive in physics at least for idealized closed systems.
Suppose we argue graphically (figure \ref{arithmetic}) that division is the sharing
of an incoming amount like a flow into $a$ parts; then why do we not follow
multiplication and ask for the answer to represent the sum inventory of all the
participants? Then one would simply say that $x/a = x$ because the amount of
information hasn't actually changed: the total
amount was conserved. Alternatively, we could introduce a partioning into new
dimensions, as in modulo arithmetic, and say that $x/a \mapsto \{
b_1,b_2,\ldots b_a\}$ for some $b_i \sim x/a$. Then we sacrifice the property of
mapping back to an object of the same or similar type, as an
inverse for multiplication within $\R1$. 

In mapping the result back into $\R1$, we make a choice about which kind of answer we want to keep.
Choosing division to be the inverse of multiplication has many advantages and few anomalies,
but the anomalies point to cases where the result makes more since in other interpretations.
Division by zero is the obvious case.
As with the debt concept for subtraction (e.g. subtracting a negative amount becoming addition 
of a positive amount), 
division by zero involves handing information across the boundary from some `interior
inventory' to or from `somewhere else'. One cannot actually realize a negative amount 
in order to subtract it, but we can {\em post hoc} interpret an addition as having the same effect. 
For equilibrium processes, however, one can take away a depletion process leading to a net shift in the balance
to a positive movement. This is like the concept of `electron holes' in solid state physics. 
In the case of division by zero, however, the handover to zero recipients is manifestly 
cavalier as it throws away the meaning of a transfer altogether. There is no recipient. In Promise Theory,
however, this does correspond to an physical scenario, where there is an
an agent offering $x$ but there being no agent accepting the amount:
\beq
x\cdot (0) : x \promise{+x} \{ \empty \}
\eeq
If an agent insists on propagating (imposing) the value, there is
no agent to catch the state, we have to define what happens to it.  The question of where the
information goes is harder to paper over with a simple concept like a
minus sign. There is now a translation into the void, from which no
inverse can recover the value.  Multiplication involves copying so it
does not conserve amounts, unless we postulate its inverse.

If we define the answer to division by the average share amount received by
each agent, then a conventional renormalization of scale would send the amount to infinity
as a limiting process, but only because we want to imagine integer divisors as discussed above.
If we ask instead what is the total inventory for all agents, thinking again about conservation,
we multiply the shares by the recipients:
\beq
x / 0 \mapsto b \cdot 0 = 0.
\eeq
Then we admit that the shares have been operated away by shrinking the
size of the receiver set to nothing, violating the conservation.
Clearly the absorbing fixed point property of zero is problematic not
so much semantically as operationally.  Once again the information is
simply cast into the void with these semantics. Which of the zeroes is
supposed to win? The existence of an agent as a container surely wins
over its imagined contents. If we have no semantic attachment there is
no rational way to resolve the difference between the zero. This might not
matter in the context of a neutral calculation, but it might be the different
between an image of a dog or a cat in a process of image reconstruction\cite{bishop1}.

What then is an appropriate return value for the function associated with
this division? We have several returnable characteristics for the process: the input
values, the dimensions of the graph, values associated with the ring
etc.  What if the return value is the number of agents involved in the
sharing? We can imagine the sharing process as being a fair weighted
split of the input $x$ into $L$ pieces.
\beq
x/L \mapsto \frac{ x}{\text{dim}(\{L\})}
\eeq
\beq
x/0 \mapsto \frac{ x}{\text{dim}(\{x\}\mapsto 0) }
\eeq
When we interpret with agents like this, we are assuming that
they are either blank ``stem cell'' agents initially or that they can
track where stuff comes from. Semantic labels make the latter possible
in computing and biology, but elementary agents may not have enough
resources to track this information.
If we think of the division as part of a process involving agents
with memory of state, then the recipient agents might already contain a non-zero amount to begin with
(like a bank account balance),
so that their final state becomes the answer rather than simply the dividing transaction.
If the agent already had a non-zero offset $b_0$ which could regularize the value $b = b_0 + x/\text{dim(\{x\})}$
it alters the way we handle the outcome.
For $\R1$ $\text{dim}(x\mapsto 0) \mapsto \infty$, but for a ring of dimension $L$, $\text{dim}(x\mapsto 0) = 1/L$,
which would imply $x/0 \mapsto x L < \infty$.

Sharing nothing between no recipients is no more or less well defined. The only natural answer
within the scope of the real numbers is 0.
The only value for which this makes invariant sense is $x=1$, because the absorbing property of 0
interacts with the idempotence of 1 as a stable fixed point under multiplication except for zero.
However, in this configuration, the division by zero eliminates the zero issue.
\beq
k \frac{x}{0} = \frac{x}{k'  0} = k \frac{1}{0} =  \mapsto 0
\eeq
This in turn suggests that, if we wish to map $0/0$ back into $\R1$ then the only plausible value is
\beq
\frac{0}{0} = 0\cdot \frac{x}{0} = 0
\eeq
The problem is not so much the value as the process semantics of the operation (see figure\ref{inversediv}).

Finally, we may note briefly that there is another example of division of states
in which a zero state can be inverted unambiguously into a finite integer number. This is for
ladder operators in Fock space algebra\cite{schweber1} or Lie algebra root and weights\cite{grouptheory}.
The zero state is defined by an absorbing state for the annihilation operator $\hat a$.
\beq
\hat a|1\rangle &=& |0\rangle\nonumber\\
\hat a|0\rangle &=& 0.
\eeq
Reconstruction, like the $\hat F$ and $\hat B$ graph operators for directed processes, give:
\beq
(a^\dagger)^n|0\rangle \mapsto |n\rangle
\eeq
It suggests that one could define
\beq
0/0 = 1, ~ ~ ~ 0/0^2 = 2, \ldots ~ ~ ~  0/0^n = n
\eeq
This is because the logarithmic property transmutes powers into addition, introducing a new
question associated with the definition of logarithms over rings and fields.
Clearly the full semantics of divisions are poorly represented by rings and fields,
in a similar way that analyticity is incompletely by the reals without a wider
scope of complex numbers.

\section{Conclusions}

Building on the Semantic Spacetime model as a set of guiding
principles for graph representations, we can simplify the selection of
proper link identification by adopting the $\gamma(3,4)$
representation to remove unnecessary ambiguities, without adopting
ontology or a detailed first order logic. The graphical properties of
the algebra are then postulated to be compatible with all expected
processes and inferences can now be written down explicitly and be
verified by matrix algebra.

It remains for future work to understand how to understand whether
knowledge is trustworthy\cite{trustnotes,trustsummary} by the balance
between the different node etcs $e,t,c$ affect the reliabilty of a
knowledge representation.  If a knowledge structure contains only
concepts, it lacks grounding in truth and can easily disconnect with
reality.  Could this explain what we are seeing with artificial
intelligence knowledge, fake news, cult beliefs and extremism?  This is a
problem of interest for a more empirical review.

Technically, the presence of absorbing states is inevitable in graphs with non
trivial adjacencies. These absorbing states need to be interpreted
carefully. If we consider the graphs to be dynamic flows, they lead to
loss of information at the edges of the system.  The wider problems of
consistent transformations within a finite structure have a common
theme: the absence of invertability, due to states that are
absorbing. In Markov processes, there is no memory to prevent loss of
information, but one assumes that information is redistributed in such
a way that it isn't actually lost. Absorbing states behave in a similar way, until
the moment we want to reverse flows that lead to them. Remedies are
analogous connection with division by zero remedies. It
 would be interesting to study the effects of the various remedies on
 matrix inversion for directed graphs in more detail.

Axioms and logical primitives already stand alone as `boundary
information' or irreducible knowledge. Such boundary information is thus
intimately related to the absorption and emissions of nodes in a
directed graph. As one uses the SST structures to scale structures,
such features will continue to exist on many scales. Whole regions of
a graph can be absorbing. Hierarchical graphs, such as taxonomies and
spanning trees, also have nodes that are both the beginning and end of
some flow.  Finiteness demands this. Only in quasi-continuum models,
groups and semi-groups, are we able to define `translationally
invariant' systems that go on and on in different directions. So the
infinity of possible outcomes for end states avoids dealing with
beginning and end states. This absence of boundary is used frequently
in physical models such as field theories, to argue for smooth
continuity.  Clearly, the concept of zero and infinity are
complementary in this process sense.

There are no moving bodies in the Semantic Spacetime discussed here,
however intrinsic properties of space can be passed along to relocate
from node to node or from agent to agent. This is called Motion Of The
Third Kind\cite{spacetime1,virtualmotion1}.  In a sense it is forced
to `invent' the semantic split between matter and spacetime in order
to resolve ambiguities about changing states, as natural philosophers
found necessary in the formative years of physics. The semantics of
material and conceptual constructs are similar. It's also reminiscent of
the arguments about whether one should call interior properties real or
not, which continues to rage in the field of Quantum Mechanics.

In modern Artificial Intelligence or reasoning models, where many of
the techniques originate from the palette of mathematical physics,
there is a tendency to paint path selection and inference with the
broad brush of probability theory, without questioning too much what
the probabilities mean.  The labels that can remember inverses have to
be trained explicitly at some expense in shadow representations of
knowledge. The relatively recent forays into context and relevance
modelling are suprisingly overdue in artificial neural network
representaions. Semantic Spacetime belongs to this effort.  Semantic
spacetime is not a natural language model.  Natural language remains a
very different way of compressing intentional descriptions with rich
semantic content, which undoubtedly relies on the capabilities of an
evolved brain for bridging the gulfs between broad and sometimes audacious inferences.
The possibility of scanning natural language and compiling a compact
SST representation is nevertheless an intriguing possibility of great
interest in connection with generative Artificial Intelligence, with
or without Large Language Models.

\bigskip
A software implementation of this work is available at \cite{sstorytime}.

\bibliographystyle{unsrt}
\bibliography{spacetime,bib} 

\begin{thebibliography}{10}

\bibitem{spacetime1}
M.~Burgess.
\newblock Spacetimes with semantics (i).
\newblock {\em arXiv:1411.5563}, 2014.

\bibitem{spacetime2}
M.~Burgess.
\newblock Spacetimes with semantics (ii).
\newblock {\em arXiv.org:1505.01716}, 2015.

\bibitem{spacetime3}
M.~Burgess.
\newblock Spacetimes with semantics (iii).
\newblock {\em arXiv:1608.02193}, 2016.

\bibitem{cognitive}
M.~Burgess.
\newblock A spacetime approach to generalized cognitive reasoning in
  multi-scale learning.
\newblock {\em arXiv:1702.04638}, 2017.

\bibitem{promisebook}
J.A. Bergstra and M.~Burgess.
\newblock {\em Promise Theory: Principles and Applications (second edition)}.
\newblock $\chi tAxis$ Press, 2014,2019.

\bibitem{rdf}
W3C.
\newblock Defining n-ary relations on the semantic web: use with individuals.
\newblock http://www.w3.org/TR/2004/WD-swbp-n-aryRelations-20040721/.

\bibitem{burgess2020testingquantitativespacetimehypothesis1}
Mark Burgess.
\newblock Testing the quantitative spacetime hypothesis using artificial
  narrative comprehension (i) : Bootstrapping meaning from episodic narrative
  viewed as a feature landscape, 2020.

\bibitem{burgess2020testingquantitativespacetimehypothesis2}
Mark Burgess.
\newblock Testing the quantitative spacetime hypothesis using artificial
  narrative comprehension (ii) : Establishing the geometry of invariant
  concepts, themes, and namespaces, 2020.

\bibitem{watt1}
D.~Watt.
\newblock {\em Programming language syntax and semantics}.
\newblock Prentice Hall, New York, 1991.

\bibitem{categories}
B.C. Pierce.
\newblock {\em Basic Category Theory for Computer Scientists}.
\newblock MIT Press, 1991.

\bibitem{quantumpictures}
B.~Coecke adn A.~Kissinger.
\newblock {\em Picturing Quantum Processes}.
\newblock Cambridge, 2017.

\bibitem{langacker1}
R.W. Langacker.
\newblock {\em Cognitive Grammar, A Basic Introduction}.
\newblock Oxford, Oxford, 2008.

\bibitem{generativesemantics}
I.~Heim and A.~Kratzer.
\newblock {\em Semantics in Generative Grammar}.
\newblock Blackwell, 1998.

\bibitem{ontologies}
J.~Strassner.
\newblock {\em Handbook of Network and System Administration}, chapter
  Knowledge Engineering Using Ontologies.
\newblock Elsevier Handbook, 2007.

\bibitem{unfolding}
G.~Deutsche.
\newblock {\em The Unfolding of Language}.
\newblock Academic Press, 2005.

\bibitem{hesse}
M.B. Hesse.
\newblock {\em Forces and Fields: The Concept of Action at a Distance in the
  History of Physics}.
\newblock Dover, 1962.

\bibitem{grimmett1}
G.R. Grimmett and D.R. Stirzaker.
\newblock {\em Probability and random processes (3rd edition)}.
\newblock Oxford scientific publications, Oxford, 2001.

\bibitem{bishop1}
C.M. Bishop and H.~Bishop.
\newblock {\em Deep Learning}.
\newblock Springer, 2023.

\bibitem{selectivestate}
N.M. Cirone, A.~Orvieto, B.~Walker, C.~Salvi, and T.~Lyons.
\newblock Theoretical foundations of deep selective state-space models, 2025.

\bibitem{bergstra6}
J.~A. Bergstra.
\newblock Division by zero, a survey of options.
\newblock {\em Transmathematica}, pages 1--20, 2019.

\bibitem{smartspacetime}
M.~Burgess.
\newblock {\em Smart Spacetime}.
\newblock $\chi$tAxis Press, 2019.

\bibitem{feynman1}
R.P. Feynman.
\newblock Space-time approach to quantum electrodynamics.
\newblock {\em Physical Review}, 76:769, 1949.

\bibitem{dyson1}
F.J. Dyson.
\newblock The radiation theories of tomonaga, schwinger and feynman.
\newblock {\em Physical Review}, 75:486, 1949.

\bibitem{abbott1}
L.F. Abbott.
\newblock Introduction to the background field method.
\newblock {\em Acta Physica Polonica}, B13:33, 1992.

\bibitem{reif1}
F.~Reif.
\newblock {\em Fundamentals of statistical mechanics}.
\newblock McGraw-Hill, Singapore, 1965.

\bibitem{myrheim1}
J.~Myrheim.
\newblock Statistical geometry.
\newblock CERN preprint TH.2538, August 1978.

\bibitem{sorkin1}
R.D. Sorkin.
\newblock Causal sers: Discrete gravity.
\newblock {\em arXiv:gr-qc/0309009}, 2003.

\bibitem{surya1}
S.~Surya.
\newblock The causal set approach to quantum gravity.
\newblock {\em arXiv:1903.11544 [gr-qc]}, 2019.

\bibitem{robertson1}
E.L. Robertson.
\newblock Triadic relations: An algebra for the semantic web.
\newblock {\em Lecture Notes in Computer Science}, 3372:91--108, 2005.

\bibitem{fanizzi2}
N.~Fanizzi, C.~d’Amato, and F.~Esposito.
\newblock Evolutionary clustering in description logics: Controlling concept
  formation and drift in ontologies.
\newblock pages 808--821, 09 2008.

\bibitem{ontologydb}
Michael Uschold.
\newblock Ontology and database schema: What’s the difference?
\newblock {\em Applied Ontology}, 10:243--258, 2015.

\bibitem{milnerbigraph}
R.~Milner.
\newblock {\em The space and motion of communicating agents}.
\newblock Cambridge, 2009.

\bibitem{burgess_search_2013}
Mark Burgess.
\newblock {\em In Search of Certainty - The Science of Our Information
  Infrastructure}.
\newblock {$\chi$tAxis} Press, November 2013.

\bibitem{stories}
A.~Couch and M.~Burgess.
\newblock Compass and direction in topic maps.
\newblock {\em (Oslo University College preprint)}, 2009.

\bibitem{inferences}
A.~Couch and M.~Burgess.
\newblock Human-understandable inference of causal relationships.
\newblock In {\em Proc. 1st International Workshop on Knowledge Management for
  Future Services and Networks}, Osaka, Japan, 2010.

\bibitem{kleinberg99authoritative}
Jon~M. Kleinberg.
\newblock Authoritative sources in a hyperlinked environment.
\newblock {\em Journal of the ACM}, 46(5):604--632, 1999.

\bibitem{graphpaper}
J.~Bjelland, M.~Burgess, G.~Canright, and K.~Engø-Monsen.
\newblock Eigenvectors of directed graphs and importance scores: dominance,
  t-rank, and sink remedies.
\newblock {\em Data Mining and Knowledge Discovery}, 20(1):98--151, 2010.

\bibitem{jan60}
M.~Burgess and A.~Couch.
\newblock On system rollback and totalized fields: An algebraic approach to
  system change.
\newblock {\em J. Log. Algebr. Program.}, 80(8):427--443, 2011.

\bibitem{shannon1}
C.E. Shannon and W.~Weaver.
\newblock {\em The Mathematical Theory of Communication}.
\newblock University of Illinois Press, Urbana, 1949.

\bibitem{cover1}
T.M. Cover and J.A. Thomas.
\newblock {\em Elements of Information Theory}.
\newblock (J.Wiley \& Sons., New York), 1991.

\bibitem{bergstra2025expressingentropycrossentropyexpansions}
Jan~A Bergstra and John~V Tucker.
\newblock Expressing entropy and cross-entropy in expansions of common meadows,
  2025.

\bibitem{bergstra2021divisionzerocommonmeadows}
Jan~A. Bergstra and Alban Ponse.
\newblock Division by zero in common meadows, 2021.

\bibitem{janderson1}
J.A.D.W. Anderson.
\newblock Perspex machine ix: Transreal analysis.
\newblock In {\em Electronic Imaging}. San Jos\', 2007.

\bibitem{janderson2}
J.A.D.W. Anderson, N.~V\"olker, and A.~A. Adams.
\newblock Perspex machine viii, axioms of transreal arithmetic.
\newblock In {\em Proc. SPIE 6499. Vision Geometry XV}, 2007.

\bibitem{bergstra2}
J.A. Bergstra and J.V. Tucker.
\newblock The rational numbers as an abstract datatype.
\newblock {\em Journal of the ACM}, 54:1--25, 2007.

\bibitem{bergstra3}
J.~A. Bergstra and J.~V. Tucker.
\newblock The data type variety of stack algebras.
\newblock {\em Annals of Pure and Applied Logic}, 73(1):11--36, 1995.

\bibitem{bergstra4}
J.~A. Bergstra and J.~V. Tucker.
\newblock Division safe calculation in totalised fields.
\newblock {\em Theory of Computing Systems}, 43:410--424, 2008.

\bibitem{bergstra5}
J.~A. Bergstra, Y.~Hirshfield, and J.~V. Tucker.
\newblock Meadows and the equational specification of division.
\newblock {\em Theoretical Computer Science}, 410:1261--1271, 2009.

\bibitem{scale1}
G.I. Barenblatt.
\newblock {\em Scaling, self-similarity, and intermediate asymptotics}.
\newblock Cambridge, 1996.

\bibitem{schweber1}
S.~Schweber.
\newblock {\em Relativitsic Quantum Field Theory}.
\newblock Harper \& Row, 1961.

\bibitem{grouptheory}
H.F. Jones.
\newblock {\em Groups, Representations, and Physics}.
\newblock Institute of Physics, 1998.

\bibitem{trustnotes}
M.~Burgess.
\newblock Notes on trust as a causal basis for social science.
\newblock {\em SSRN Archive, available at
  http://dx.doi.org/10.2139/ssrn.4252501 (DOI: 10.2139/ssrn.4252501)}, August
  2022.

\bibitem{trustsummary}
M.~Burgess.
\newblock Trust and trustability: An idealized operational theory of economic
  attentiveness.
\newblock {\em preprint paper (DOI: 10.13140/RG.2.2.26862.28480/1)}, April
  2023.

\bibitem{virtualmotion1}
M.~Burgess.
\newblock Motion of the third kind (i) notes on the causal structure of virtual
  processes for privileged observers.
\newblock DOI: 10.13140/RG.2.2.30483.35361 (notes available on Researchgate),
  2021.

\bibitem{sstorytime}
M.~Burgess.
\newblock Sstorytime project.
\newblock \url{https://github.com/markburgess/SSTorytime/tree/main}, 2025.

\end{thebibliography}

\appendix
\section{Examples}

Only events can lead to events.
\begin{enumerate}
\item Hammering leads to noise

 Possible: ``The activity of hammering'' L(leads to) ``the event of noise''
 Wrong: ``The activity of hammering'' C(contains) ``noise''

 Possible: ``The concept of hammering'' E(may have property) ``noise''

\item His stubbornness led to indignation

 Possible: ``The event in which he was stubborn'' (led to) ``An event in which there was indignation''

 Wrong: ``The concept of stubbornness'' E(has the property) ``the concept of indignation''

\item That cake is just like your house!

 Unlikely: ``The cake (thing)'' N(is similar to) ``your house (thing)''

 Likely: ``The appearance of the cake (concept)''  E(has property of mapping to) ``the appearance of your house''

\item The virus caused his death

 Impossible: ``The virus concept or thing'' L(led to) ``the event of his death'' 

 Possible: ``the viral infection event'' L(led to)  ``the event of his death''

\item Professor Plumb murders Ms Scarlet in the library

 ``The event of plumb murders scarlet'' (is an example of) ``concept of murder''

 ``The event of plumb murders scarlet'' E(has the attribute) ``concept of murder''

 ``The concept of plumb murders scarlet'' E(is an example of) ``concept of murder''

\end{enumerate}
\end{document}